\documentclass[preprint,number,review,12pt]{elsarticle} %CON DOPPIA INTERLINEA

\usepackage[top=2.5cm,bottom=2.5cm,left=2.5cm,right=2.5cm]{geometry}
\usepackage{amssymb}
\usepackage{xcolor}
\usepackage{graphicx}
\graphicspath{ {./images/} }
\usepackage{multirow} 
\usepackage{amsmath} \usepackage{amsfonts}
\usepackage{textcomp}
\usepackage{dblfloatfix}
\usepackage{caption}
\usepackage{cleveref}
\usepackage{subcaption}
\usepackage{lineno}
\DeclareMathAlphabet{\mathpzc}{OT1}{pzc}{m}{it}
\usepackage{amsmath}
\usepackage{bbm}
\usepackage[mathscr]{euscript}
\usepackage{gensymb}
\usepackage{color}
\usepackage{enumitem} % Pacote para personalizar listas
\usepackage{ragged2e} % Para justificar o texto
\usepackage{overpic}

\newcommand{\Expect}{{\rm I\kern-.3em E}}

\begin{document}

\begin{frontmatter}



%\title{Surrogate Modeling of Multiple Engineering Design Parameters for Optimization in Performance-Based Earthquake Engineering}

% \title{Metamodel-Based Framework for Stochastic Simulations in Performance-Based Earthquake Engineering Optimization}

% ADDRESSING STATISTICAL UNCERTAINTY IN A SURROGATE MODEL FRAMEWORK

%\title{A Stratified Stochastic Emulation Framework for Performance-Based Risk Optimization of Structures under Seismic Hazard}

\title{Stochastic Emulation using Generalized Stratified Sampling for Performance-Based Risk Optimization of Structures}


\author[inst1]{Isabela D. Rodrigues}

\author[inst2]{Seymour M.J. Spence\corref{cor1}}
\ead{smjs@umich.edu}
\cortext[cor1]{Corresponding author}

\author[inst3]{Henrique M. Kroetz}

\author[inst1]{Andr\'{e} T. Beck}

\affiliation[inst1]{organization={Department of Structural Engineering},%Department and Organization
            addressline={São Carlos School of Engineering, University of São Paulo}, 
            city={São Carlos},
            postcode={13566-590}, 
            state={São Paulo},
            country={Brazil}}

\affiliation[inst2]{organization={Department of Civil and Environmental Engineering},%Department and Organization
            addressline={University of Michigan}, 
            city={Ann Arbor},
            postcode={48109}, 
            state={Michigan},
            country={USA}}

\affiliation[inst3]{organization={Center for Marine Studies},%Department and Organization
            addressline={Federal University of Paraná}, 
            city={Pontal do Paraná},
            postcode={13566-590}, 
            state={Paraná},
            country={Brazil}}

\begin{abstract}

Metamodels are instrumental in reducing the computational burden associated with nested reliability analyses and optimization loops in Performance-Based Risk Optimization (PBRO) of structures under stochastic loads. In this context, stochastic emulators are particularly useful because they approximate response distributions while accounting for the intrinsic stochasticity of the simulator. Among these methods, Stochastic Polynomial Chaos Expansion (SPCE) is especially attractive because it does not require replications of nonlinear analyses at fixed input conditions. However, SPCE may present limitations in accurately representing extreme responses in the tails of structural response distributions. To address this limitation, this study proposes a framework that combines Generalized Stratified Sampling (GSS) with SPCE. The GSS scheme partitions the input space into strata according to the intensity of the hazard, improving the representation of extreme responses, while independent SPCE emulators are trained within each stratum. The conditional exceedance probabilities estimated in each stratum are then recombined using the total probability theorem to evaluate the probabilistic constraints in the optimization problem. The proposed GSS-SPCE framework is applied to the optimal design of the cross-sectional areas of buckling-restrained braces in a two-story steel building. The objective is to minimize the initial construction cost while satisfying prescribed probabilistic performance constraints. In the case study, the strata are defined based on the spectral acceleration associated with the seismic excitation, while the design variables are included as additional inputs to the SPCE emulators. This allows the response distributions associated with each design candidate to be efficiently and accurately estimated. Results show that the proposed framework accurately estimates structural response distributions, including their tail regions, while substantially reducing the number of nonlinear model evaluations required for PBRO.

\end{abstract}


\end{frontmatter}


\section{Introduction}

The quantification of structural performance in terms of risk and reliability has become increasingly common in the assessment and design of structures exposed to natural hazards, including earthquakes \cite{krawinkler1999challenges} and extreme winds \cite{spence2022performance}. By explicitly accounting for uncertainties in hazard intensity and structural parameters, risk- and reliability-based performance assessments provide information that helps designers and stakeholders make more informed decisions regarding structural design, assessment, and management, with the aim of reducing economic losses and other adverse consequences. In this context, structural performance assessment can be integrated naturally into optimization frameworks, in which design alternatives are systematically explored to achieve an appropriate balance among structural safety, functionality, and economy. This class of problems has been referred to in the literature as Performance-Based Design Optimization (PBDO) \cite{hassanzadeh2024performance}; in the present study, following \citet{rodrigues2026incorporating}, the term Performance-Based Risk Optimization (PBRO) is adopted.

The PBRO framework can be interpreted within the broader context of Reliability-Based Design Optimization (RBDO) and Risk Optimization (RO) \cite{beck2012comparison}. In both formulations, reliability analysis is nested within the optimization and must be repeated throughout the search for the optimum. This nested structure introduces a significant computational burden, which may become prohibitive for high-dimensional problems or complex structural systems \cite{spence2014performance,rastegaran2022multi,NI2024103707,lopez2025efficient,movaghar2025}. Among the various alternatives proposed to address this issue, metamodels (or surrogate models) have shown significant promise in approximating the input--output relationships of high-fidelity numerical models at substantially reduced computational cost \cite{spence2018optimization,kim2021reliability,miguel2023performance,miguel2024reliability,dos2024sequential,GU2023103514,misra2025metaimnet}.

Although metamodels provide an efficient data-driven approach, their development for PBRO problems is further complicated by the stochastic nature of seismic and wind excitations. The estimation of performance metrics typically relies on repeated nonlinear simulations to obtain structural response quantities, commonly referred to as Engineering Demand Parameters (EDPs), such as displacements and accelerations. Because the response depends on the random process representing the excitation, its evaluation becomes particularly demanding within an optimization framework, where each iteration requires response estimates for the current design candidate. Representative PBRO studies using metamodels include \citet{gidaris2015performance}, who incorporated Kriging into an optimization framework to approximate repair costs and reference velocities from stochastic simulations as functions of the design variables. Other studies have approximated the conditional distributions of EDPs. \citet{gidaris2015kriging} developed a Kriging metamodel to approximate the median and standard deviation of the logarithmic structural response for use in fragility and risk calculations. An alternative formulation was later proposed by \citet{kyprioti2021kriging}, in which the EDP distribution was approximated directly using Kriging, thereby reducing the number of replications of the nonlinear simulation. \citet{kim2024active} approximated the moments of the logarithmic structural response using a heteroscedastic Gaussian process metamodel within an active-learning-based RBDO framework. More recently, \citet{rodrigues2026incorporating} proposed a noisy Kriging framework trained to predict the median, dispersion, and correlation of the EDP distribution, while incorporating statistical uncertainty arising from a limited number of ground motions through a noise term. Further methodologies for RBDO problems under stochastic excitation were reviewed by \citet{jerez2022reliability}.

Despite the efficiency of the aforementioned metamodeling frameworks, their application still faces challenges related to the direct propagation of high-dimensional uncertainties through nonlinear simulations. These challenges have motivated the recent development of stochastic emulators, which aim to approximate entire response distributions rather than deterministic model outputs, with applications involving seismic \cite{zhu2023seismic,yi2024stochastic,yi2025multi,kim2025uncertainty} and wind excitations \cite{zhu2020replication,KROETZ2026122515,MACEDO2026103971}. In this context, a nonlinear structural model subjected to stochastic excitation can be interpreted as a stochastic simulator: for the same set of input parameters, different realizations of the stochastic excitation may lead to different structural response values. As a result, the structural response is a random output for fixed inputs, thereby accounting for the variability induced by the stochastic excitation. Recent formulations of stochastic emulators include, among others, Generalized Lambda Models (GLMs) \cite{zhu2021emulation,zhu2021global} and Stochastic Polynomial Chaos Expansions (SPCEs) \cite{zhu2023stochastic}.

In this paper, SPCE is employed as the stochastic emulator, following its recent application to reliability analysis problems \cite{pires2025reliability,pires2025alspce}. SPCE introduces an artificial latent variable to represent the intrinsic stochasticity of the simulator, together with a noise term that improves numerical stability. This formulation allows the emulator to approximate general, nonparametric conditional response distributions without requiring replications of the original simulator at the same input point \cite{zhu2023stochastic}. More recently, \citet{moustapha2026high} incorporated SPCE into an RBDO framework to approximate conditional response distributions and compared its performance with that of Kriging metamodels. However, as pointed out by \citet{zhu2023seismic}, SPCE still faces limitations in seismic fragility analysis, particularly when failure data are scarce at large damage thresholds, because it is not specifically designed to accurately represent the tail of the EDP distribution. To address this limitation, \citet{zhu2023seismic} suggested using advanced sampling schemes to better represent extreme events, and this direction is explored in the present work.

Advanced sampling schemes, such as importance sampling \cite{melchers1989importance,au2003important} and subset simulation \cite{au2001estimation,au2003subset}, have been widely studied and applied in reliability analysis to reduce the large number of model evaluations required to estimate small failure probabilities using Monte Carlo methods \cite{wang2019hamiltonian,chen2022riemannian,ARUNACHALAM2023102310,xian2024relaxation,LEE2025110947}. These schemes generate samples that more effectively explore the failure region, thereby increasing the proportion of realizations that contribute to the failure probability estimate. Several studies have also combined advanced sampling schemes with metamodels to further reduce the computational burden associated with reliability evaluations involving very small failure probabilities \cite{DUBOURG201347,ZHU2020106644,XIAO2020113336,YU2023108722}.

With this background, this work combines the Generalized Stratified Sampling (GSS) scheme, formally introduced by \citet{arunachalam2023generalized}, with SPCE to improve the approximation of the tails of conditional EDP distributions within a Performance-Based Engineering (PBE) framework. The proposed framework, referred to as GSS-SPCE, is subsequently used to solve a PBRO problem. In this framework, GSS partitions the input space according to hazard intensity, improving the representation of rare events, while separate SPCE emulators are trained within each stratum with the design variables included as inputs. The main contribution of this work is the integration of an advanced sampling scheme with stochastic emulation to improve tail representation and enable the efficient estimation of probabilistic constraints within PBRO.

The remainder of the paper is organized as follows. Section \ref{sec:setting} presents the problem definition, including the PBE and PBRO concepts adopted in this work. Section \ref{sec:methods} provides background on the stochastic simulation, SPCE, and GSS methods underlying the proposed framework. Section \ref{sec:proposed} presents the proposed GSS-SPCE framework and its implementation within PBRO. Section \ref{sec:casestudy} demonstrates the framework through a case study involving the optimal design of the cross-sectional areas of Buckling-Restrained Braces (BRBs) in a two-story steel building, with the objective of minimizing the initial construction cost while satisfying prescribed probabilistic performance constraints.

% ---------------------
% ---------------------

\section{Problem definition}
\label{sec:setting}

\subsection{Assessment of structural performance}
\label{subsec:strucperf}

The assessment of building performance generally requires the characterization of a structural response quantity, represented by the random variable $Y$, under uncertainties in the system parameters and load excitation. For a given response level $y$, the exceedance probability is defined as:
\begin{equation}
p\left(y\right)=P\left(Y>y\right)=1-F_Y\left(y\right)
\label{EqStruPerfm1}
\end{equation}
where $p\left(y\right)$ is the probability that the structural response $Y$ exceeds the response level $y$, and $F_Y$ is the cumulative distribution function (CDF) of $Y$. Structural performance can then be assessed by evaluating this probability at a prescribed critical value $y_{\mathrm{cr}}$, which gives rise to the design constraint:
\begin{equation}
p\left(y_{\mathrm{cr}}\right)=P\left(Y>y_{\mathrm{cr}}\right)\leq p_{\mathrm{t}}
\label{EqStruPerfm1_2}
\end{equation}
where $p_{\mathrm{t}}$ is the acceptable exceedance probability specified by the designer. Therefore, the following probabilistic integral must be evaluated:
\begin{equation}
p\left(y_{\mathrm{cr}}\right)=1-F_Y\left(y_{\mathrm{cr}}\right)
=1-\int_{\Omega_{y_{\mathrm{cr}}}}f_{\mathbf{X}}\left(\mathbf{x}\right)d\mathbf{x}
\label{EqStruPerfm2}
\end{equation}
where $\mathbf{X}$ is a random vector defined over the input space $\mathcal{D}_{\mathbf{X}}\subseteq\mathbb{R}^{n_{\mathrm{RV}}}$ and represents the uncertainties associated with the system parameters and load excitation, $n_{\mathrm{RV}}$ is the number of random variables, $f_{\mathbf{X}}$ is the joint probability density function (PDF) of $\mathbf{X}$, and $\Omega_{y_{\mathrm{cr}}}=\{\mathbf{x}:Y(\mathbf{x})\leq y_{\mathrm{cr}}\}$ is the domain in which the structural response does not exceed the critical value.

When the stochastic nature of the load excitation is explicitly modeled, as in this work, the integral in Eq.~(\ref{EqStruPerfm2}) becomes high-dimensional and generally must be estimated using simulation methods \cite{spence2018optimization}. For instance, if Monte Carlo simulation is considered, the following steps are required to obtain $F_Y\left(y_{\mathrm{cr}}\right)$, considering the structural response as a nonlinear function $y=g_{\mathrm{NL}}\left(\mathbf{x}\right)$:
\begin{enumerate}[noitemsep,nolistsep]
\item Generate a realization $\mathbf{x}^{(i)}$ of the uncertain vector $\mathbf{X}$;
\item Generate a realization of the stochastic load using the appropriate components of $\mathbf{x}^{(i)}$;
\item Evaluate the corresponding realization of the structural response, $y^{(i)}$, using $g_{\mathrm{NL}}$;
\item If the target total number of samples has been generated, proceed to the next step. Otherwise, return to Step 1;
\item Estimate $F_Y\left(y_{\mathrm{cr}}\right)$ from the samples of $Y$.
\end{enumerate}
In Step 5, $F_Y\left(y_{\mathrm{cr}}\right)$ can be estimated directly from the samples using a nonparametric approach or by fitting a parametric probability distribution.

If the acceptable exceedance probability is small, a large number of system evaluations, $n_{\mathrm{sys}}$, must be performed. For example, under crude Monte Carlo simulation, if $p_{\mathrm{t}}=10^{-k}$, approximately $n_{\mathrm{sys}}=10^{k+2}$ model evaluations are required to estimate $p\left(y_{\mathrm{cr}}\right)$ with a coefficient of variation of $10\%$. For computationally intensive nonlinear models, this requirement is generally prohibitive \cite{arunachalam2023generalized}.

% -----------------------
% -----------------------

\subsection{Optimization problem}
\label{subsec:otproblem}

Considering the structural performance assessment defined in Section~\ref{subsec:strucperf}, the design of a system can be formulated as a PBRO problem. The optimization consists of selecting values for a set of parameters, denoted by $\mathbf{d}$, that satisfy a set of performance constraints while minimizing the system cost $C$. The vector $\mathbf{d}$ may represent structural element sizes, while the performance constraints may be associated with the requirements defined in Eq.~(\ref{EqStruPerfm1_2}). The optimization problem can be written as:
\begin{equation}
\begin{aligned}
& \textrm{Find} \quad \mathbf{d} = \{d_1,\ldots,d_{N_d}\}^{T} \\
& \textrm{to minimize} \quad C(\mathbf{d}) \\
& \textrm{subject to} \quad P\left(Y_j\left(\mathbf{d},\mathbf{X}\right)>y_{\mathrm{cr},j}\right)\leq p_{\mathrm{t},j}
\quad \textrm{for } j=1,\ldots,N_c \\
& \mathbf{d}\in\mathcal{D} \\
& \mathbf{X}\in\mathcal{D}_{\mathbf{X}}
\end{aligned}
\label{EqOt1}
\end{equation}
where $\mathbf{d}$ is an $N_d$-dimensional vector of design variables belonging to the design space $\mathcal{D}\subseteq\mathbb{R}^{N_d}$, $Y_j\left(\mathbf{d},\mathbf{X}\right)$ is the $j$th structural response of interest, $y_{\mathrm{cr},j}$ is the corresponding critical value, $p_{\mathrm{t},j}$ is the associated acceptable exceedance probability, and $N_c$ is the number of constraints.

Solving the optimization problem is not straightforward due to the reliability analyses nested within the optimization loop. Since the constraints are implicit functions of $\mathbf{d}$, their evaluation requires a full probabilistic assessment, as described in Section~\ref{subsec:strucperf}, for each design candidate. This computational burden limits the use of evolutionary optimizers, while the implicit dependence of the constraints on $\mathbf{d}$ also makes gradient-based optimization challenging due to the difficulty of evaluating constraint gradients. Although metamodeling schemes have been proposed to approximate the structural response, limitations remain when high-dimensional uncertainties must be propagated directly through nonlinear simulations and when the relevant tail probabilities, as in Eq.~(\ref{EqStruPerfm1}), are very small.

To address these limitations, this work proposes combining an advanced sampling scheme with a stochastic emulation approach to accelerate the evaluation of the probabilistic constraints for any given design $\mathbf{d}$. The GSS-SPCE method uses SPCE to emulate the structural response distributions associated with the constraints, while GSS partitions the input space according to hazard intensity to improve the approximation of the response distribution tails. The proposed approach facilitates the efficient solution of the PBRO problem defined in Eq.~(\ref{EqOt1}) by allowing the probabilistic constraints to be evaluated rapidly without additional calls to the nonlinear structural model.

% ---------------------------
% ---------------------------
% ---------------------------

\section{Background methods}
\label{sec:methods}

The implementation of the GSS-SPCE framework requires the structural performance assessment to be formulated as a stochastic simulation problem. This formulation enables the structural response distribution to be emulated using SPCE, while the input space is partitioned through the GSS scheme. The background formulations of these methods are presented in the following sections.

\subsection{Stochastic simulators}
\label{subsec:StochasticSimulation}

To formulate the nonlinear structural response model $y=g_{\mathrm{NL}}(\mathbf{x})$ as a stochastic simulator, $Y_{\mathbf{x}}=g_{\mathrm{NL}}\left(\mathbf{x};\omega\right)\equiv\mathcal{M}_s\left(\mathbf{x},\omega\right)$, the uncertainty associated with stochastic excitation can be separated into two main components. The first component is described by parametric explanatory variables that characterize the hazard intensity, such as seismic spectral acceleration or wind velocity. These variables are commonly referred to as intensity measures (IMs) and are explicitly included in the input vector $\mathbf{x}$. The second component is associated with the complex physical mechanisms involved in the generation and propagation of the stochastic excitation. In seismic applications, this component is commonly referred to as record-to-record (RTR) variability. It represents the different possible load time histories that may occur for the same IM value, limiting the extent to which the excitation can be completely described by the selected parametric explanatory variables \cite{yi2025multi,baker2008uncertainty}.

This separation naturally leads to the formulation of the structural response as a stochastic simulator. While a deterministic simulator produces a unique scalar- or vector-valued output for a fixed input vector, a stochastic simulator produces random outputs even when the explicit input parameters are fixed. In this context, the output randomness represents the intrinsic variability of the simulator, which may account, for instance, for the effects of RTR variability.

A stochastic simulator can be written as:
\begin{equation}
\begin{aligned}
\mathcal{M}_s:\mathcal{D}_{\mathbf{X}}\times\Omega &\rightarrow \mathbb{R}\\
\left(\mathbf{x},\omega\right) &\mapsto \mathcal{M}_s\left(\mathbf{x},\omega\right)
\end{aligned}
\label{EqStocSim1}
\end{equation}
where $\Omega$ is the sample space associated with the internal stochasticity of the simulator and $\omega\in\Omega$. Therefore, for a fixed input vector $\mathbf{x}$, different realizations of $\omega$ may lead to different values of the structural response.

For a fixed $\mathbf{x}\in\mathcal{D}_{\mathbf{X}}$, the response $Y_{\mathbf{x}}=\mathcal{M}_s\left(\mathbf{x},\cdot\right):\Omega\rightarrow\mathbb{R}$ is therefore a random variable. Accordingly, when the input vector is fixed at $\mathbf{x}^{(0)}$, each simulator run corresponds to a different event $\omega^{(j)}$ and produces a realization $\mathcal{M}_s\left(\mathbf{x}^{(0)},\omega^{(j)}\right)$.

This behavior is common in structural engineering applications involving stochastic loads. For ground motions, the response may remain random even when event parameters such as magnitude, source-to-site distance, and site conditions are fixed, due to RTR variability \cite{zhu2023seismic,yi2025multi}. Similarly, under wind loads, the structural response may vary due to the turbulent component of the excitation, even for fixed IM values \cite{pires2025alspce,KROETZ2026122515}.

% -------------------

\subsection{Stochastic Polynomial Chaos Expansion}
\label{subsec:SPCE}

SPCE is a stochastic emulator designed to reproduce the conditional distribution of the response $Y$ of a stochastic simulator given the input vector $\mathbf{X}$, that is, $Y_{\mathbf{x}}\equiv Y\mid\mathbf{X}=\mathbf{x}$. Originally introduced by \citet{zhu2023stochastic}, SPCE extends the framework of Polynomial Chaos Expansions (PCEs) to represent the intrinsic stochasticity of the original simulator $\mathcal{M}_s$.

PCEs constitute a well-established and widely used metamodeling technique for deterministic simulators. Considering the random input vector $\mathbf{X}\in\mathbb{R}^{n_{\mathrm{RV}}}$ described by the joint PDF $f_{\mathbf{X}}$, a deterministic model output $y=\mathcal{M}_d(\mathbf{x})$ can be approximated by an orthogonal polynomial expansion, expressed as:
\begin{equation}
\mathcal{M}_d\left(\mathbf{x}\right)
\approx
\sum_{\boldsymbol{\alpha}\in\mathcal{A}}
c_{\boldsymbol{\alpha}}\,
\psi_{\boldsymbol{\alpha}}\!\left(\mathbf{x}\right)
\label{Eq_PCE1}
\end{equation}
where $\psi_{\boldsymbol{\alpha}}$ denotes a multivariate polynomial basis function that is orthonormal with respect to the joint probability density function $f_{\mathbf{X}}$, $\boldsymbol{\alpha}=\left[\alpha_1,\alpha_2,\ldots,\alpha_{n_{\mathrm{RV}}}\right]$ is the corresponding multi-index, and $c_{\boldsymbol{\alpha}}$ is the associated expansion coefficient to be estimated. The set $\mathcal{A}\subset\mathbb{N}^{n_{\mathrm{RV}}}$ is a finite set of multi-indices obtained by applying a truncation rule. The standard truncation scheme consists of selecting all polynomials whose total degree is less than or equal to a prescribed value $p_{\mathrm{dg}}$, such that $\mathcal{A}^{p_{\mathrm{dg}},n_{\mathrm{RV}}}=\left\{\boldsymbol{\alpha}\in\mathbb{N}^{n_{\mathrm{RV}}}:\sum_{j=1}^{n_{\mathrm{RV}}}\alpha_j\leq p_{\mathrm{dg}}\right\}$.

To construct the SPCE, a latent random variable $Z$ is introduced to represent the intrinsic stochasticity of the stochastic simulator. Let $F_{Y\mid\mathbf{X}}\left(y\mid\mathbf{x}\right)$ denote the conditional CDF of the response for a fixed input vector $\mathbf{x}$. Using the probability integral transform and assuming a prescribed distribution for the latent variable $Z$, the conditional response can be expressed as $Y_{\mathbf{x}}=F_{Y\mid\mathbf{X}}^{-1}\left(F_Z\left(Z\right)\mid\mathbf{x}\right)$, where $F_Z$ is the CDF of $Z$. Assuming that $Y$ has finite variance, $F_{Y\mid\mathbf{X}}^{-1}\left(F_Z\left(Z\right)\mid\mathbf{x}\right)$ can be represented by a PCE in the augmented input space $\left(\mathbf{X},Z\right)$. By adding a noise term $\epsilon\sim\mathcal{N}(0,\sigma^2)$ to improve numerical stability, the SPCE can be written as:
\begin{equation}
Y_{\mathbf{x}}\approx\widetilde{Y}_{\mathbf{x}}
=
\sum_{\boldsymbol{\alpha}\in\mathcal{A}}
c_{\boldsymbol{\alpha}}\,
\psi_{\boldsymbol{\alpha}}\!\left(\mathbf{x},Z\right)
+\epsilon
\label{Eq_SPCE1}
\end{equation}
The SPCE defined in Eq.~(\ref{Eq_SPCE1}) is constructed by estimating the coefficients $c_{\boldsymbol{\alpha}}$ and the standard deviation of the noise term, $\sigma$. In this work, the calibration follows the procedure described in \cite{zhu2023stochastic,luthen2024uqlab}. This procedure also includes the adaptive selection of the latent-variable distribution and the truncation scheme $\mathcal{A}\subset\mathbb{N}^{n_{\mathrm{RV}}+1}$. For this purpose, a Design of Experiments (DoE) is defined as:
\begin{equation}
\begin{aligned}
\mathcal{X}_S &=
\left[\mathbf{x}^{(1)},\ldots,\mathbf{x}^{(S)}\right]\\
\mathcal{Y}_S &=
\left[
\mathcal{M}_s\left(\mathbf{x}^{(1)},\omega^{(1)}\right),
\ldots,
\mathcal{M}_s\left(\mathbf{x}^{(S)},\omega^{(S)}\right)
\right]
\end{aligned}
\label{Eq_SPCE2}
\end{equation}
where the matrix $\mathcal{X}_S$ of support points contains $S$ realizations of $\mathbf{X}$ sampled in $\mathcal{D}_{\mathbf{X}}$, and $\mathcal{Y}_S$ contains the corresponding stochastic simulator evaluations. Each support point is evaluated only once.

As discussed in the Introduction, SPCE is particularly suitable for the present problem because it emulates the conditional response distribution without imposing a predefined parametric form and without requiring replications of the original simulator at the same input point. Once the SPCE is built, its realizations can be used to estimate response distributions and exceedance probabilities, as defined in Eq.~(\ref{EqStruPerfm2}). Nevertheless, the estimation of small exceedance probabilities requires an accurate representation of the distribution tails, which may be challenging when rare response levels are poorly represented in the training set \cite{zhu2023seismic}. For this reason, SPCE is combined in this work with the GSS scheme. The GSS procedure is used to generate stratum-specific DoE samples and the corresponding model evaluations, which are then used to train independent SPCEs within each stratum.

% ----------------------

\subsection{Generalized Stratified Sampling scheme}
\label{subsec:SS}

The GSS scheme adopted in this work follows the methodologies proposed by \citet{ARUNACHALAM2023102310} and \citet{arunachalam2023generalized} for reliability assessment problems involving natural hazards and the estimation of multiple small failure probabilities. In the present study, GSS is used to generate the DoE samples within each stratum required for the calibration of SPCE. Accordingly, only Phase~I of the generalized stratified sampling framework proposed by \citet{arunachalam2023generalized} is considered, since the objective is to construct stratum-specific samples rather than to estimate stratum-specific failure probabilities directly.

The basic idea is to partition the input space $\mathcal{D}_{\mathbf{X}}$ into a collection of $n_{\mathbb{S}}$ disjoint strata:
\begin{equation}
\left\{\mathbb{S}_i\right\}_{i=1}^{n_{\mathbb{S}}}
\label{EqPF1}
\end{equation}
such that:
\begin{equation}
\bigcup_{i=1}^{n_{\mathbb{S}}}\mathbb{S}_i
=
\mathcal{D}_{\mathbf{X}}
\quad\text{and}\quad
\mathbb{S}_i\cap\mathbb{S}_j
=
\varnothing
\quad\text{for}\quad
i\neq j
\label{EqPF2}
\end{equation}
The partition is defined in terms of a scalar stratification variable $\mathcal{V}$, which may be either an input random variable or the output of an auxiliary computational model $\mathcal{H}$ that depends on a subset of the input uncertainties. In the latter case, $\mathcal{V}$ is evaluated as $\mathcal{V}=\mathcal{H}(\boldsymbol{\tau})$, where $\boldsymbol{\tau}$ denotes the subset of input components required to compute the stratification variable. The use of $\mathcal{H}$ is particularly convenient when its computational cost is much lower than that of the nonlinear structural model $g_{\mathrm{NL}}$, allowing a large number $\hat{n}$ of realizations of $\mathcal{V}$ to be generated.

After selecting the number of strata $n_{\mathbb{S}}$ and the probability constant $p_{\mathrm{GSS}}$, the stratum boundaries are defined adaptively from the empirical distribution of $\mathcal{V}$. Following \citet{arunachalam2023generalized}, values of $p_{\mathrm{GSS}}$ commonly lie in the range $[0.1,0.3]$. Equivalently, if the $\hat{n}$ realizations of $\mathcal{V}$ are sorted in increasing order, $\mathcal{V}_i$ is taken as the $\left(1-p_{\mathrm{GSS}}^i\right)\hat{n}$th ordered value for $i=1,\ldots,n_{\mathbb{S}}-1$. Setting $\mathcal{V}_0=-\infty$ and $\mathcal{V}_{n_{\mathbb{S}}}=\infty$, the strata are induced in the input space by:
\begin{equation}
\mathbb{S}_i
=
\left\{
\boldsymbol{\tau}:
\mathcal{V}\in\left(\mathcal{V}_{i-1},\mathcal{V}_i\right]
\right\},
\quad
i=1,\ldots,n_{\mathbb{S}}
\label{EqPF3}
\end{equation}
Once the strata are defined, samples are generated within each stratum to form the DoE. The components of $\boldsymbol{\tau}$ are sampled conditionally on the corresponding stratum, while the remaining input variables are sampled according to their prescribed distributions.

When the stratification variable and the stratum boundaries are appropriately defined, GSS provides a more efficient representation of rare-event regions than standard Monte Carlo sampling using the same number of model evaluations. This motivates its use in the present work to improve the coverage of the input space used to calibrate the SPCE, particularly in regions associated with the tails of the structural response distribution.

% -------------------
% -------------------
% -------------------

\section{Proposed framework}
\label{sec:proposed}

\subsection{Overview}
\label{subsec:Overview}

The GSS-SPCE framework proposed in this work is used to emulate the distribution of the structural response generated by the stochastic simulator associated with the nonlinear structural model and the stochastic load excitation. Instead of approximating only a deterministic response value for a given input vector $\mathbf{x}$, the proposed framework generates realizations of the conditional emulated response $\widetilde{Y}_{\mathbf{x}}$ following the SPCE formulation defined in Eq.~(\ref{Eq_SPCE1}). By evaluating the trained stratum-specific SPCEs over samples of $\mathbf{X}$, the resulting realizations are combined to estimate the unconditional CDF of the structural response, denoted by $F_{\widetilde{Y}}$, and the corresponding exceedance probability:
\begin{equation}
\widetilde{p}\left(y\right)
=
P\left(\widetilde{Y}>y\right)
\label{EqPF4}
\end{equation}
as an approximation of the performance metric defined in Eq.~(\ref{EqStruPerfm1}).

Once the stochastic emulators are trained and their accuracy is assessed in representing the unconditional response distributions, they are used within the optimization framework defined in Eq.~(\ref{EqOt1}) to estimate the probabilistic constraints for each design candidate.

The following subsections describe the construction of the GSS-SPCE in the augmented input space and its subsequent use in the optimization problem.

\subsection{Construction of the SPCE in the augmented input space}
\label{subsec:SPCEAugmented}

The complete set of uncertainties involved in the structural performance assessment can be conceptually divided into model-related uncertainties, $\mathbf{X}_m$, load-intensity uncertainties, $\mathbf{X}_h$, and the stochastic variability of the load excitation, $\mathbf{X}_w$. These sources may be represented by $\boldsymbol{\Xi}=\left[\mathbf{X}_m^T,\mathbf{X}_h^T,\mathbf{X}_w^T\right]^T$.

As defined in Eq.~(\ref{Eq_SPCE1}), the SPCE is formulated in the augmented input space $\left(\mathbf{X},Z\right)$, where $\mathbf{X}=\left[\mathbf{X}_m^T,\mathbf{X}_h^T\right]^T$ contains the variables explicitly sampled and provided as inputs to the emulator. The latent random variable $Z$ represents the remaining stochastic variability associated with $\mathbf{X}_w$, which is not explicitly included in $\mathbf{X}$. To avoid repeated nonlinear-model evaluations when estimating the probabilistic constraints within the optimization, the design variables $\mathbf{d}$ are also included as emulator inputs. Therefore, the GSS-SPCE is constructed in the fully augmented input space $\left(\mathbf{U},Z\right)$, where $\mathbf{U}=\left[\mathbf{d}^T,\mathbf{X}_m^T,\mathbf{X}_h^T\right]^T$. For the purpose of DoE generation and SPCE training, the design variables are sampled from uniform distributions over the design domain $\mathcal{D}$. As a result, the explicit input vector of the stochastic emulator is $\mathbf{U}$, while $Z$ remains a latent variable. For a realization $\mathbf{u}=\left[\mathbf{d}^T,\mathbf{x}^T\right]^T$, the structural response estimated by the SPCE is conditioned on the design variables and the explicitly sampled random inputs and is denoted by $\widetilde{Y}_{\mathbf{u}}=\widetilde{Y}\left(\mathbf{d},\mathbf{x}\right)$. For a fixed design vector $\mathbf{d}^{(\ell)}\in\mathcal{D}$ within the optimization process, random realizations of $\mathbf{X}$ are propagated through the trained SPCEs to generate samples of $\widetilde{Y}\left(\mathbf{d}^{(\ell)},\mathbf{X}\right)$.

Alternative formulations are also possible. For instance, part of the uncertainty associated with the explicitly sampled random inputs $\mathbf{X}$ could instead be represented in the latent space through $Z$. In the limiting case, $Z$ would represent all stochastic variability associated with $\boldsymbol{\Xi}$, leaving only the design variables as explicit emulator inputs. The resulting mapping from the design variables to the structural response would still constitute a stochastic simulator, with the latent variable accounting for all output randomness.

%The outputs of the SPCE are schematically represented in Fig.~\ref{SPCE_outputs}.
%
%\begin{figure}[htb!]
%\centering
%\includegraphics[scale=0.25]{Figures_R0/Figure1.pdf}
%\caption{Representation of the SPCE outputs in the augmented space.}
%\label{SPCE_outputs}
%\end{figure}

% --------------------------
% --------------------------

\subsection{Stratified scheme for constructing the GSS-SPCE}
\label{subsec:GSSSPCEscheme}

Following the GSS formulation described in Section~\ref{subsec:SS}, the stratification variable is evaluated from the subset $\boldsymbol{\tau}$ of the hazard-related variables $\mathbf{X}_h$. Once the strata $\{\mathbb{S}_i\}_{i=1}^{n_{\mathbb{S}}}$ are defined, samples of $\mathbf{X}_h$ are selected within each stratum by conditioning the components $\boldsymbol{\tau}$ on the corresponding stratum. If $P\left(\mathbb{S}_{n_{\mathbb{S}}}\right)\approx10^{-k}$, approximately $10^{k+2}$ evaluations of $\mathcal{H}$ are required to obtain an expected $10^2$ samples in the last stratum, corresponding approximately to a coefficient of variation of $10\%$ in the estimated stratum probability \cite{arunachalam2023generalized}.

The remaining components of $\mathbf{X}_h$, if any, are sampled according to their prescribed distributions. These samples are then combined with uniformly distributed realizations of the design variables $\mathbf{d}$ and realizations of the model-related uncertainties $\mathbf{X}_m$ to form the stratum-specific input vectors $\mathbf{U}^{(i)}$ used to train the SPCEs. This leads to a stratum-specific emulated response, denoted by $\widetilde{Y}^{(i)}$. The exceedance probability in Eq.~(\ref{EqPF4}) can then be expressed within each stratum as:
\begin{equation}
\widetilde{p}^{(i)}\left(y\right)
=
P\left(
\widetilde{Y}^{(i)}\left(\mathbf{U}^{(i)}\right)>y
\mid
\mathbb{S}_i
\right)
\label{EqPF5}
\end{equation}
Following the law of total probability, the unconditional exceedance probability is obtained by recombining the stratum-specific conditional probabilities as:
\begin{equation}
\widetilde{p}\left(y\right)
=
\sum_{i=1}^{n_{\mathbb{S}}}
\widetilde{p}^{(i)}\left(y\right)
P\left(\mathbb{S}_i\right)
\label{EqPF6}
\end{equation}
where $P\left(\mathbb{S}_i\right)$ denotes the probability of occurrence of the $i$th stratum. This recombination is valid because the strata $\{\mathbb{S}_i\}_{i=1}^{n_{\mathbb{S}}}$ form a mutually exclusive and collectively exhaustive partition of the input space, as defined in Eq.~(\ref{EqPF2}).

In the general SPCE formulation, the DoE was denoted by $\mathcal{X}_S$ in Eq.~(\ref{Eq_SPCE2}) and defined in terms of the input vector $\mathbf{X}$. In the proposed GSS-SPCE framework, the corresponding DoE is defined in the augmented input space and denoted by $\mathcal{U}_S^{(i)}$ for each stratum $\mathbb{S}_i$ as:
\begin{equation}
\begin{aligned}
\mathcal{U}_S^{(i)}
&=
\left[
\mathbf{u}^{(1,i)},
\ldots,
\mathbf{u}^{(S,i)}
\right]
\\
\mathbf{u}^{(j,i)}
&=
\left[
\left(\mathbf{d}^{(j,i)}\right)^T,
\left(\mathbf{x}_m^{(j,i)}\right)^T,
\left(\mathbf{x}_h^{(j,i)}\right)^T
\right]^T,
\qquad
j=1,\ldots,S
\end{aligned}
\label{Eq_DoE_GSSSPCE}
\end{equation}
where $j$ indexes the support points and $\mathbf{x}_h^{(j,i)}$ denotes a realization of the hazard-related variables associated with stratum $\mathbb{S}_i$.

% -------------------
% -------------------

\subsection{Algorithm}
\label{subsec:Algorithm}

\begin{figure*}
\centering
\includegraphics[scale=0.5]{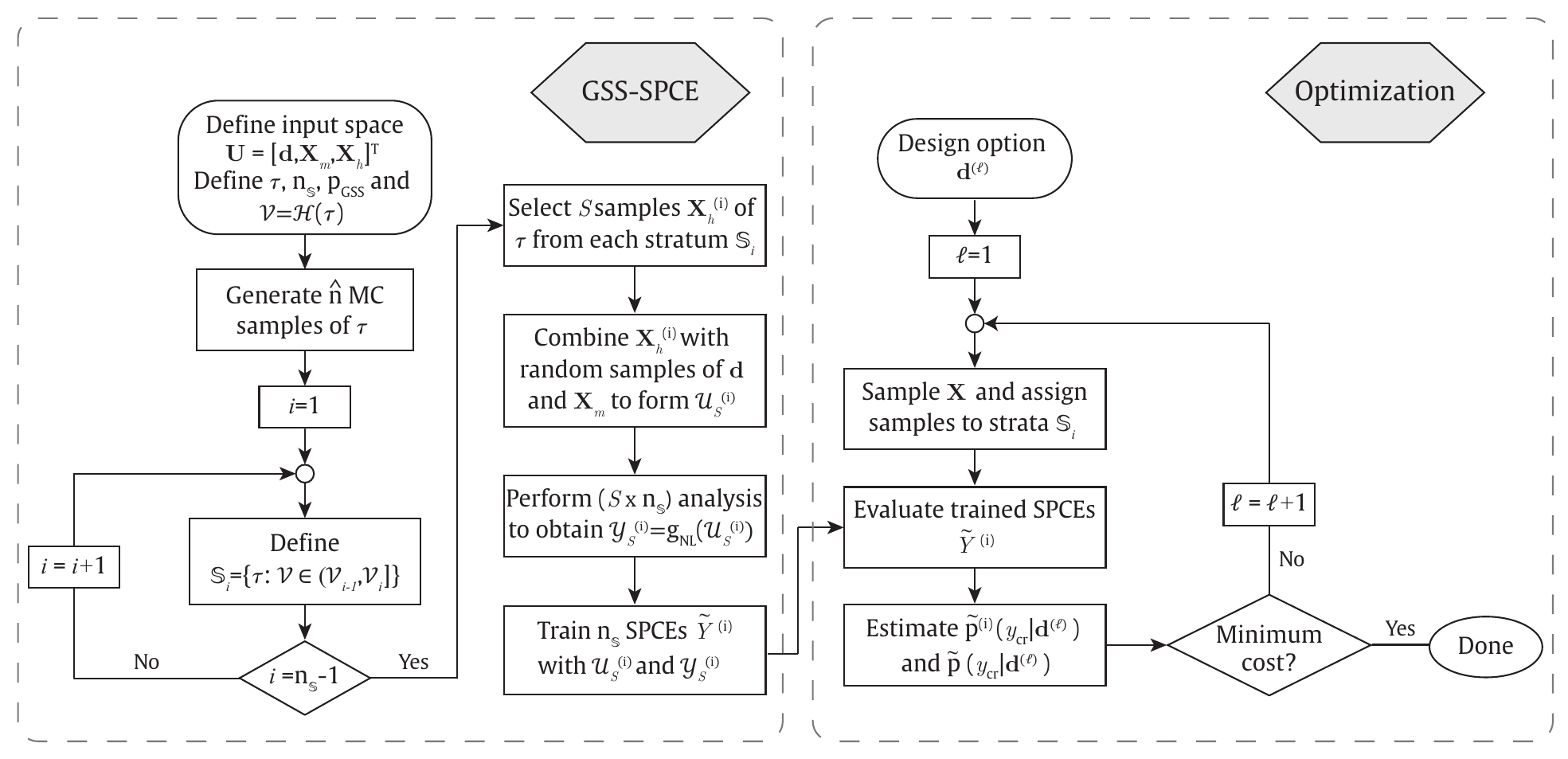}
\caption{Flowchart of the proposed methodology.}
\label{Fig_Algorithm}
\end{figure*}

The overall procedure of the proposed framework is summarized in the flowchart shown in Fig.~\ref{Fig_Algorithm}. The following steps are implemented to partition the input space, select samples within each stratum, train the SPCE emulators, and estimate the exceedance probabilities within the optimization loop:

\begin{enumerate}[noitemsep,nolistsep]

    \item Define the augmented input vector $\mathbf{U}=\left[\mathbf{d}^{T},\mathbf{X}_m^{T},\mathbf{X}_h^{T}\right]^{T}$.

    \item Define the subset $\boldsymbol{\tau}$ of the hazard-related variables $\mathbf{X}_h$ used to evaluate the stratification variable $\mathcal{V}=\mathcal{H}(\boldsymbol{\tau})$, together with the number of strata $n_{\mathbb{S}}$ and the probability constant $p_{\mathrm{GSS}}\in[0.1,0.3]$. Generate a sufficiently large number $\hat{n}$ of realizations of $\boldsymbol{\tau}$ to define the stratum boundaries adaptively, as described in Section~\ref{subsec:SS}. This procedure partitions the input space into $n_{\mathbb{S}}$ disjoint strata $\{\mathbb{S}_i\}_{i=1}^{n_{\mathbb{S}}}$.

    \item For each stratum $\mathbb{S}_i$, select $S$ realizations of the hazard-related variables, denoted by $\{\mathbf{x}_h^{(j,i)}\}_{j=1}^{S}$. Combine these samples with realizations of the design variables $\{\mathbf{d}^{(j,i)}\}_{j=1}^{S}$ and the model-related uncertainties $\{\mathbf{x}_m^{(j,i)}\}_{j=1}^{S}$ to form the stratum-specific DoE $\mathcal{U}_S^{(i)}$ in the augmented input space.

    \item Perform $n_{\mathbb{S}}\times S$ evaluations of the stochastic simulator $\mathcal{M}_s$ to obtain the corresponding response vectors $\mathcal{Y}_S^{(i)}$ for $i=1,\ldots,n_{\mathbb{S}}$.

    \item Train one SPCE for each stratum. Each emulator $\widetilde{Y}^{(i)}$ is calibrated using the stratum-specific DoE $\mathcal{U}_S^{(i)}$ and the corresponding response vector $\mathcal{Y}_S^{(i)}$.

    \item Within the optimization loop, for each design candidate $\mathbf{d}^{(\ell)}$, generate a sufficiently large number of realizations of $\mathbf{X}$. For each realization, identify the corresponding stratum $\mathbb{S}_i$ from the value of $\boldsymbol{\tau}$ and evaluate the associated emulator $\widetilde{Y}^{(i)}\left(\mathbf{d}^{(\ell)},\mathbf{x}_m,\mathbf{x}_h\right)$. Estimate the exceedance probability conditioned on each stratum, $\widetilde{p}^{(i)}\left(y\mid\mathbf{d}^{(\ell)}\right)$, using Eq.~(\ref{EqPF5}), and recombine the stratum-specific probabilities using Eq.~(\ref{EqPF6}) to obtain the unconditional exceedance probability $\widetilde{p}\left(y\mid\mathbf{d}^{(\ell)}\right)$.

    \item Repeat Step~6 for the successive design candidates generated by the optimization algorithm until its termination criteria are satisfied and the optimal design $\mathbf{d}^{\star}$ is obtained by minimizing the cost function while satisfying the probabilistic constraints.

\end{enumerate}

% -------------------
% -------------------
% -------------------

\section{Case study}
\label{sec:casestudy}

\subsection{Overview}
\label{sec:CSoverview}

\begin{figure*}[htb!]
\centering
\begin{subfigure}{0.4\textwidth}
\centering
\includegraphics[width=\linewidth]{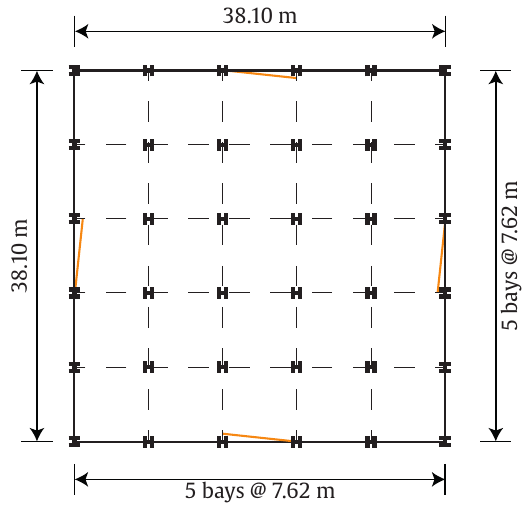}
\caption{}
\label{Fig1a_GSS_SPCE}
\end{subfigure}
\hfill
\begin{subfigure}{0.4\textwidth}
\centering
\includegraphics[width=\linewidth]{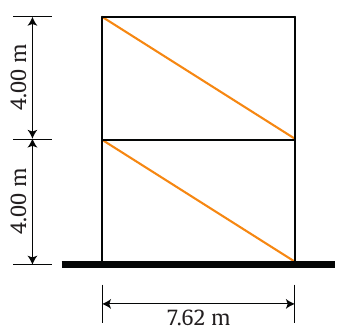}
\caption{}
\label{Fig1b_GSS_SPCE}
\end{subfigure}
\caption{Case-study structure: (a) typical plan view of the building; (b) elevation of the steel BRBF.}
\label{Fig1_GSS_SPCE}
\end{figure*}

The proposed framework is demonstrated through a case study involving a two-story buckling-restrained braced-frame (BRBF) archetype building designed in accordance with \cite{american2002seismic,american2013minimum}. Buckling-restrained braces (BRBs) are concentric structural elements designed to prevent global brace buckling and concentrate inelastic deformations within a yielding steel core \cite{uang2004research}. This mechanism enhances lateral stiffness and energy dissipation capacity, thereby improving the seismic performance of the building by limiting damage to the surrounding structural members. The selected archetype is documented in the National Institute of Standards and Technology (NIST) GCR 10-917-8 report \cite{nist2010evaluation} and is assumed to be located in urban California. Plan and elevation views of the building are presented in Figs.~\ref{Fig1a_GSS_SPCE} and~\ref{Fig1b_GSS_SPCE}, respectively.

The building performance in this case study is assessed with respect to its response under seismic excitation. The seismic inputs are generated using the stochastic ground-motion model described in \ref{sec:sample:SGMM}. The optimization problem considers the cross-sectional areas of the BRBs on the first and second stories, denoted by $d_1$ and $d_2$, respectively, as design variables. The objective is to determine the design vector $\mathbf{d}$ that minimizes the construction cost of the frame while satisfying prescribed probabilistic performance constraints. The PBRO problem is formulated as:
\begin{equation}
\begin{aligned}
\textrm{Find} \quad
&\mathbf{d}
=
\left[d_1,d_2\right]^T
\\
\textrm{to minimize} \quad
&C(\mathbf{d})
\\
\textrm{subject to} \quad
&\widetilde{p}_1\left(y_{\mathrm{cr}}\mid\mathbf{d}\right)
=
P\left(
\widetilde{Y}_1\left(\mathbf{d},\mathbf{X}\right)
>
y_{\mathrm{cr}}
\right)
\leq
p_{\mathrm{t}}
\\
&
\widetilde{p}_2\left(y_{\mathrm{cr}}\mid\mathbf{d}\right)
=
P\left(
\widetilde{Y}_2\left(\mathbf{d},\mathbf{X}\right)
>
y_{\mathrm{cr}}
\right)
\leq
p_{\mathrm{t}}
\\
&
20\leq d_1\leq70\,\mathrm{cm}^2
\\
&
20\leq d_2\leq70\,\mathrm{cm}^2
\end{aligned}
\label{EqCaseStudy}
\end{equation}
where $\widetilde{Y}_1\left(\mathbf{d},\mathbf{X}\right)$ and $\widetilde{Y}_2\left(\mathbf{d},\mathbf{X}\right)$ denote the emulated first- and second-story interstory drifts, respectively, while $\widetilde{p}_1\left(y_{\mathrm{cr}}\mid\mathbf{d}\right)$ and $\widetilde{p}_2\left(y_{\mathrm{cr}}\mid\mathbf{d}\right)$ are their corresponding exceedance probabilities. The optimal design obtained by solving Eq.~(\ref{EqCaseStudy}) is denoted by $\mathbf{d}^{\star}$.

For demonstration purposes, the critical interstory drift threshold and acceptable exceedance probability are set to $y_{\mathrm{cr}}=15\%$ and $p_{\mathrm{t}}=4\times10^{-3}$, respectively.

The building is located on Site Class C and has a typical story height of $4.0\,\mathrm{m}$. The beams and columns are made of ASTM A992 Grade 50 steel, while the BRBs are made of ASTM A36 steel. Additional design details can be found in \cite{nist2010evaluation}. Table~\ref{GSSSPCETable1} summarizes the geometric specifications of the structural elements.

\begin{table}[]
\caption{Member sizes for the BRBF.}
\centering
\small
{\tabcolsep8pt
\begin{tabular}{c c c}
\hline
Story & Beam size & Column size \\
\hline
1 & W21$\times$73 & W14$\times$82 \\
2 & W21$\times$50 & W14$\times$82 \\
\hline
\end{tabular}}
\label{GSSSPCETable1}
\end{table}

The construction cost of the steel frame, denoted by $C(\mathbf{d})$ in Eq.~(\ref{EqCaseStudy}), is estimated as:
\begin{equation}
C(\mathbf{d})
=
C_{\mathrm{s}}
\left(
W_{\mathrm{b}}+W_{\mathrm{c}}
\right)
+
C_{\mathrm{BRB}}W_{\mathrm{BRB}}
\label{EqCaseStudy2}
\end{equation}
where $C_{\mathrm{s}}$ is the cost per unit weight of structural steel, $C_{\mathrm{BRB}}$ is the cost per unit weight of the BRBs, and $W_{\mathrm{b}}$, $W_{\mathrm{c}}$, and $W_{\mathrm{BRB}}$ are the total weights of the beams, columns, and BRBs, respectively. The unit costs adopted in this study are $C_{\mathrm{s}}=\$4.72/\mathrm{kg}$ \cite{ghasemof2022multi} and $C_{\mathrm{BRB}}=\$8.99/\mathrm{kg}$ \cite{guerrero2017evaluation}. The nominal weights of the beams and columns are obtained from the American Institute of Steel Construction (AISC) database, version 16.0 \cite{design1999specification}. The total weight of the BRBs is calculated as $W_{\mathrm{BRB}}=\rho\sum_{b=1}^{N_{\mathrm{BRB}}}A_bL_b$, where $N_{\mathrm{BRB}}$ is the total number of BRBs, and $A_b$ and $L_b$ are the cross-sectional area and length of the $b$th BRB, respectively. The material density is taken as $\rho=7850\,\mathrm{kg/m^3}$ for ASTM A36 steel.

\subsection{Finite element model of the structural system}
\label{subsec:FEmodel}

A two-dimensional nonlinear dynamic finite element (FE) model of the two-story BRBF archetype is developed in OpenSees \cite{mckenna2006opensees}. The model represents one exterior braced frame in the direction of analysis. Only the bare steel components of the BRBF are explicitly modeled; the composite floor slabs and gravity framing system are not directly represented. The floor slabs are assumed to act as rigid diaphragms, such that the seismic mass tributary to the analyzed frame is lumped at the floor levels.

Beams and columns are modeled using elastic beam-column elements with concentrated flexural springs at their ends to represent the plastic hinge regions. The nonlinear response of these springs is defined through the moment--rotation relationships for each member section. The cyclic behavior of the steel springs is represented using the \textit{Steel02} material model available in OpenSees, originally based on the Giuffr\`e--Menegotto--Pinto formulation with isotropic hardening parameters \cite{filippou1983effects}. The rotational springs are implemented using \textit{zeroLength} elements. A Young's modulus of $200\,\mathrm{GPa}$ and a yield strength of $345\,\mathrm{MPa}$ are adopted for the beams and columns, together with a post-yield strain-hardening ratio of $1\%$. The elastic portions of the beams and columns between the end springs are represented by \textit{elasticBeamColumn} elements using the corresponding cross-sectional properties.

The BRBs are modeled using \textit{twoNodeLink} elements with axial response only. The cyclic behavior of the BRB steel core is also represented using the \textit{Steel02} material model. A yield strength of $290\,\mathrm{MPa}$ and a post-yield strain-hardening ratio of $2\%$ are adopted for the BRB core.

Rayleigh damping is assigned to reproduce a damping ratio of $2\%$ in the first and third vibration modes. Geometric nonlinearities are considered by explicitly including $P$--$\Delta$ effects in the beam and column elements. Gravity loads are applied at the floor levels consistently with the tributary vertical loads of the exterior braced bay, while the seismic masses are assigned according to the tributary floor area associated with the analyzed frame under the rigid-diaphragm assumption. The gravity-load and seismic-mass assumptions are defined based on the archetype design information provided in \cite{nist2010evaluation}.

% ---------------------
% ---------------------

\subsection{GSS-SPCE implementation}
\label{subsec:GSSSPCEimplementation}

\subsubsection{Input space and probability models}
\label{subsubsec:preamble}

To emulate the structural response $\widetilde{Y}$ and estimate the exceedance probability defined in Eq.~(\ref{EqPF4}), the input space must first be defined. %No model-related random variables $\mathbf{X}_m$ are considered in this case study. Therefore, the explicit input vector of the GSS-SPCE is given by:
In this example, the input vector of the GSS-SPCE is given by:
\begin{equation}
\mathbf{U}
=
\left[
\mathbf{d}^{T},
\mathbf{X}_h^{T}
\right]^{T}
\label{EqCaseStudyInput}
\end{equation}
where $\mathbf{d}=\left[d_1,d_2\right]^T$ is the vector of design variables defined in Eq.~(\ref{EqCaseStudy}), and $\mathbf{X}_h=\left[M_w,r\right]^T$ is the vector of hazard-related random variables. Here, $M_w$ is the earthquake moment magnitude and $r$ is the epicentral distance. The remaining stochastic variability associated with the ground-motion excitation is represented implicitly through the latent variable $Z$ of the SPCE. For the construction of the DoE, the design variables are sampled uniformly within their admissible ranges, while the hazard-related variables are sampled according to their prescribed probability distributions.

The earthquake magnitude $M_w$ is modeled using a truncated Gutenberg--Richter distribution \cite{arunachalam2023generalized}, with PDF given by:
\begin{equation}
f_{M_w}\left(M_w\right)
=
\frac{
\beta
\exp\left[
-\beta\left(
M_w-M_{w,\min}
\right)
\right]
}{
1-
\exp\left[
-\beta\left(
M_{w,\max}-M_{w,\min}
\right)
\right]
}
\label{EqSGMM3}
\end{equation}
where $M_{w,\min}=6$, $M_{w,\max}=8$, and $\beta=0.9\ln\left(10\right)$.

The epicentral distance $r$ is modeled as a lognormal random variable with a median of $5\,\mathrm{km}$ and a coefficient of variation of $0.4$. For the construction of the DoE, the BRB areas $d_1$ and $d_2$ are sampled independently from uniform distributions over their admissible ranges, such that $d_1,d_2\in\left[20,70\right]\,\mathrm{cm}^2$. The number of support points within each stratum is set to $S=2000$, with each support point corresponding to one realization of the explicit input vector $\mathbf{U}$. These support points form the stratum-specific DoE: $\mathcal{U}_S^{(i)}$.

% -------------------

\subsubsection{Stratum-specific samples}
\label{subsubsec:stratawisesamples}

To construct the stratum-specific DoEs, the number of strata is set to $n_{\mathbb{S}}=5$, and the probability constant is set to $p_{\mathrm{GSS}}=0.2$. The stratification variable $\mathcal{V}$ is chosen as the spectral acceleration at the first-mode period, $S_a(\bar{T})$, considering a damping ratio of $5\%$. This choice is based on \citet{arunachalam2023generalized}, who observed an increasing trend in drift ratios with increasing spectral acceleration, thereby supporting the use of $S_a(\bar{T})$ as an effective stratification variable.

Accordingly, the auxiliary computational model $\mathcal{H}$ used to evaluate $\mathcal{V}$ consists of the stochastic ground-motion model, parameterized by the hazard-related variables, followed by the evaluation of the response of a linear oscillator. Thus, $\mathcal{V}=S_a(\bar{T})=\mathcal{H}(\boldsymbol{\tau})$, where $\boldsymbol{\tau}=\mathbf{X}_h=\left[M_w,r\right]^T$ in this case study. The reference period $\bar{T}=0.624\,\mathrm{s}$ is computed for the design $d_1=d_2=45\,\mathrm{cm}^2$, corresponding to the midpoint of each admissible design-variable range, and is used throughout the stratification process.

As discussed in Section~\ref{subsec:GSSSPCEscheme}, a region with a probability of the order of $10^{-3}$ requires approximately $10^5$ evaluations of $\mathcal{H}$ to obtain an expected $10^2$ samples within that region. In this study, $\hat{n}=10^7$ realizations of $\boldsymbol{\tau}$ are generated using Monte Carlo simulation to provide a sufficiently large candidate pool for selecting $S=2000$ support points within the final stratum.

\begin{figure*}[]
\centering
\begin{subfigure}{0.46\textwidth}
\centering
\includegraphics[scale=0.28]{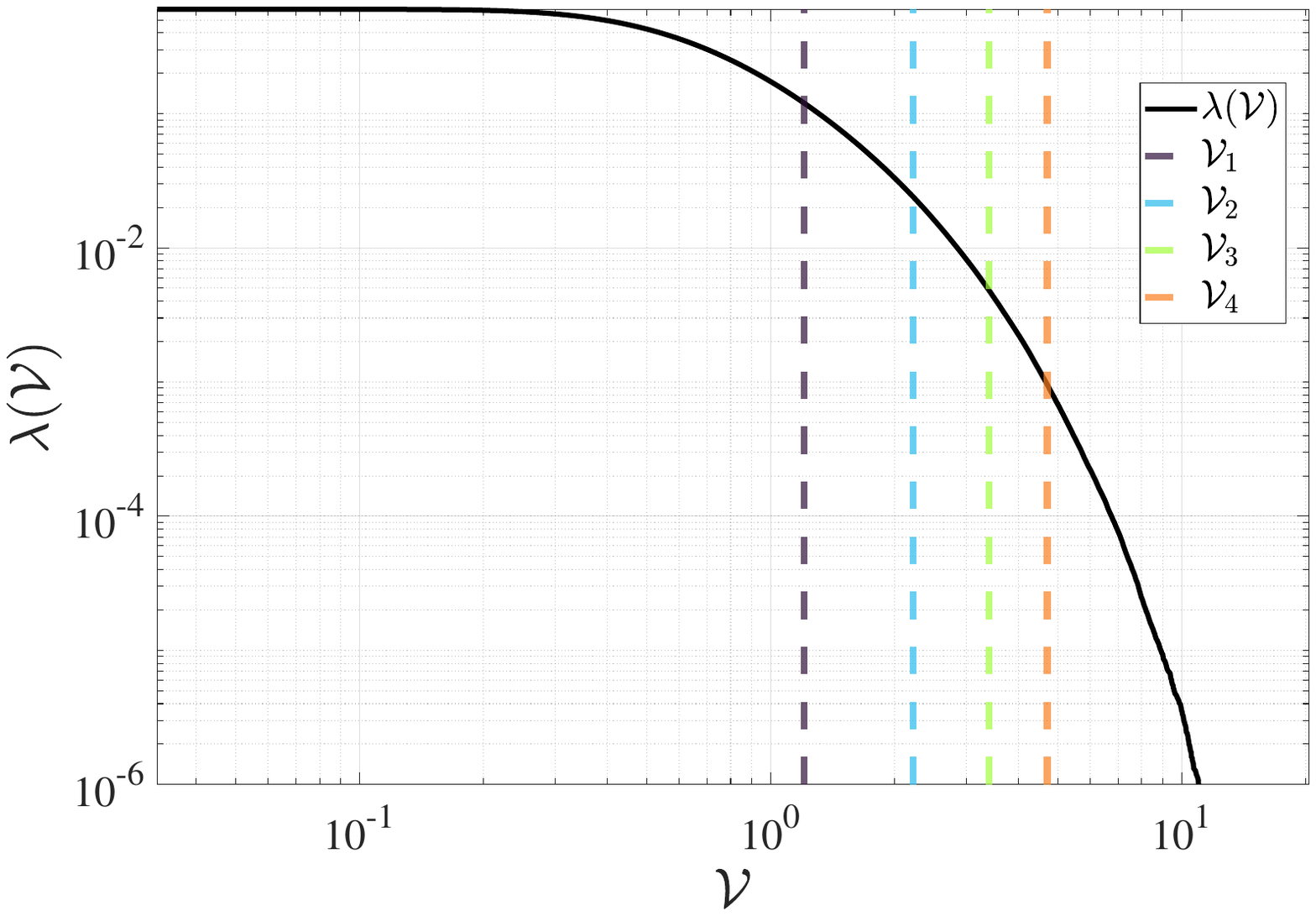}
\caption{ } 
\label{Fig_StrataWiseSamples_a} 
\end{subfigure}
\begin{subfigure}{0.46\textwidth}
\centering
\includegraphics[scale=0.28]{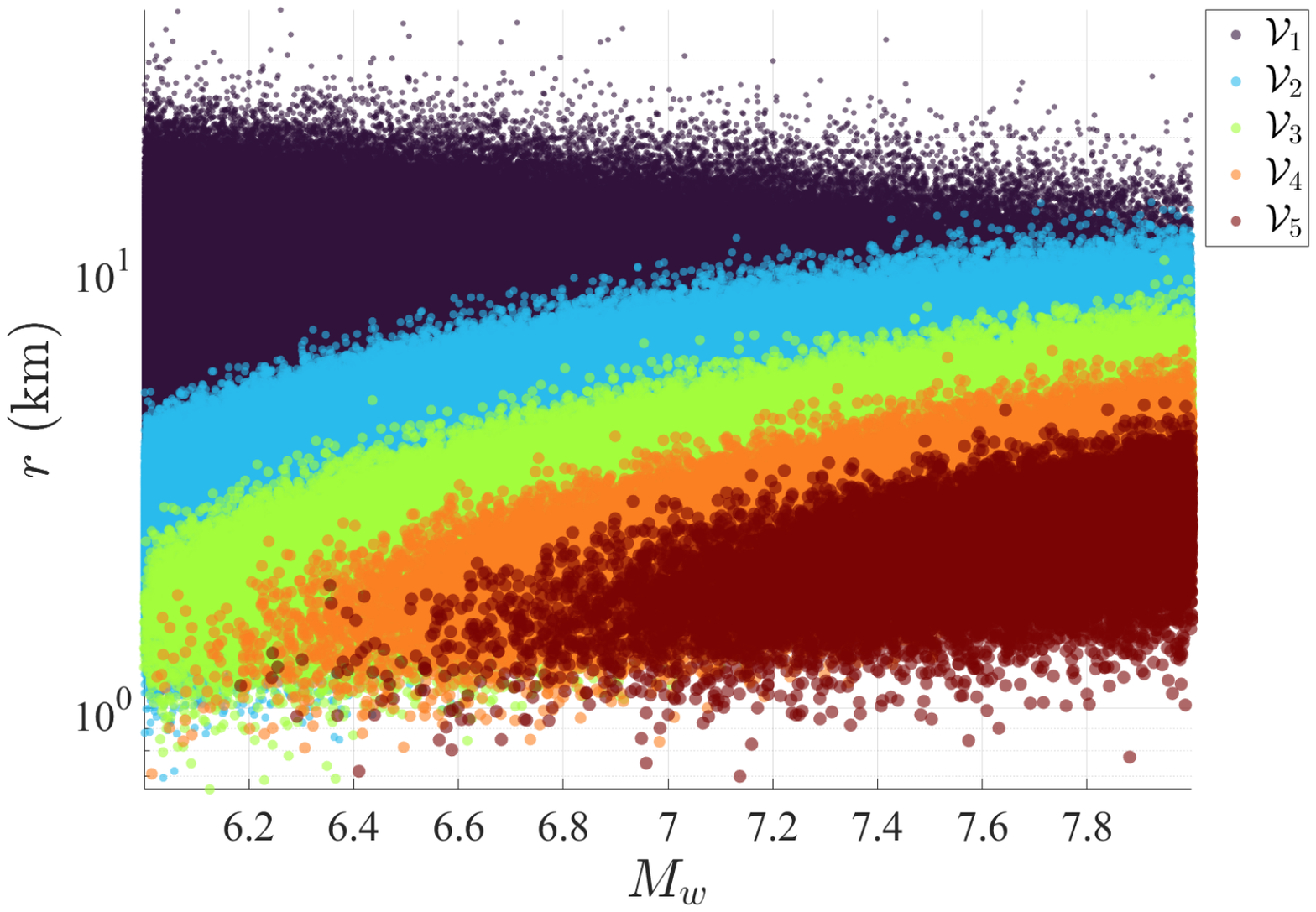}
\caption{ } 
\label{Fig_StrataWiseSamples_b} 
\end{subfigure}
\caption{Stratum-specific samples: (a) estimated empirical spectral-acceleration hazard curve $\lambda(\mathcal{V})$ and the corresponding stratum boundaries; (b) stratum-specific scatter of $M_w$ and $r$.}
\label{Fig_StrataWiseSamples}
\end{figure*}

Because spectral acceleration is nonnegative, the lower boundary is set to $\mathcal{V}_0=0$. The resulting stratum boundaries are $\left\{\mathcal{V}_0,\mathcal{V}_1,\mathcal{V}_2,\mathcal{V}_3,\mathcal{V}_4,\mathcal{V}_5\right\}=\left\{0,1.207,2.223,3.404,4.710,\infty\right\}$ in units of $g$. The estimated empirical spectral-acceleration hazard curve and the corresponding stratum boundaries are shown in Fig.~\ref{Fig_StrataWiseSamples_a}, assuming $\lambda_{M_w6}=0.6$ as the occurrence rate associated with the lower magnitude threshold $M_{w,\min}=6$. Figure~\ref{Fig_StrataWiseSamples_b} presents the stratum-specific scatter of the seismic hazard parameters $M_w$ and $r$.

\newpage

% ----------------------
% ------------------------

\subsection{Results}
\label{subsec:results}

The results are presented in the following subsections: (1) construction and validation of the GSS-SPCE over the augmented input space $\mathbf{U}$; and (2) optimization results obtained using the trained GSS-SPCEs.

% ----------------------

\subsubsection{GSS-SPCE results}
\label{subsubsec:gssspceresults}

\begin{figure}[]
\centering
\includegraphics[scale=0.29]{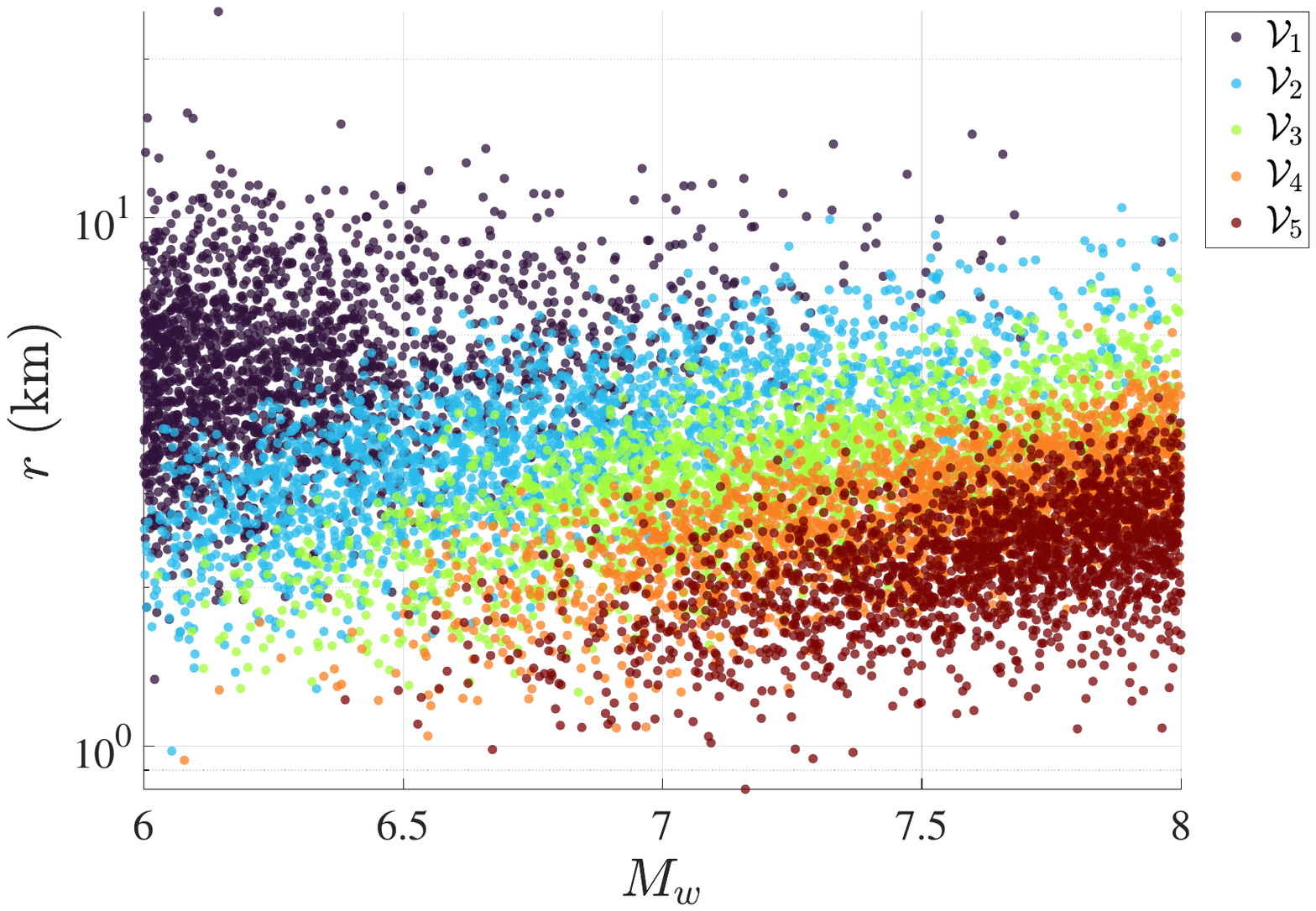}
\caption{Selected samples of $M_w$ and $r$ for the $S=2000$ support points within each stratum.}
\label{Fig_GSS_SPCE_selectedsamples}
\end{figure}

From the candidate samples generated in Section~\ref{subsubsec:stratawisesamples}, $S=2000$ realizations of $M_w$ and $r$ are randomly selected within each stratum to serve as hazard-related support points for the construction of the SPCEs. The selected samples within each stratum are shown in Fig.~\ref{Fig_GSS_SPCE_selectedsamples}.

For each stratum, $S$ support points of the design vector $\mathbf{d}$ are generated using Latin hypercube sampling and combined with the hazard-related support points to form the stratum-specific DoE: $\mathcal{U}_S^{(i)}$. The accuracy of the GSS-SPCE is first assessed within each stratum. For this purpose, an independent reference dataset containing 5000 realizations of the seismic hazard variables $M_w$ and $r$ is generated for each stratum. The design variables $d_1$ and $d_2$ are also sampled within their admissible ranges to form the corresponding input vectors $\mathbf{U}^{(i)}$. For each input vector, a stochastic ground-motion realization is generated and a nonlinear time-history analysis (NLTHA) is performed using the stochastic simulator $\mathcal{M}_s$ to obtain the reference response samples $Y_{\mathrm{R}}^{(i)}$.

\begin{figure*}[]
\centering
\includegraphics[scale=0.4]{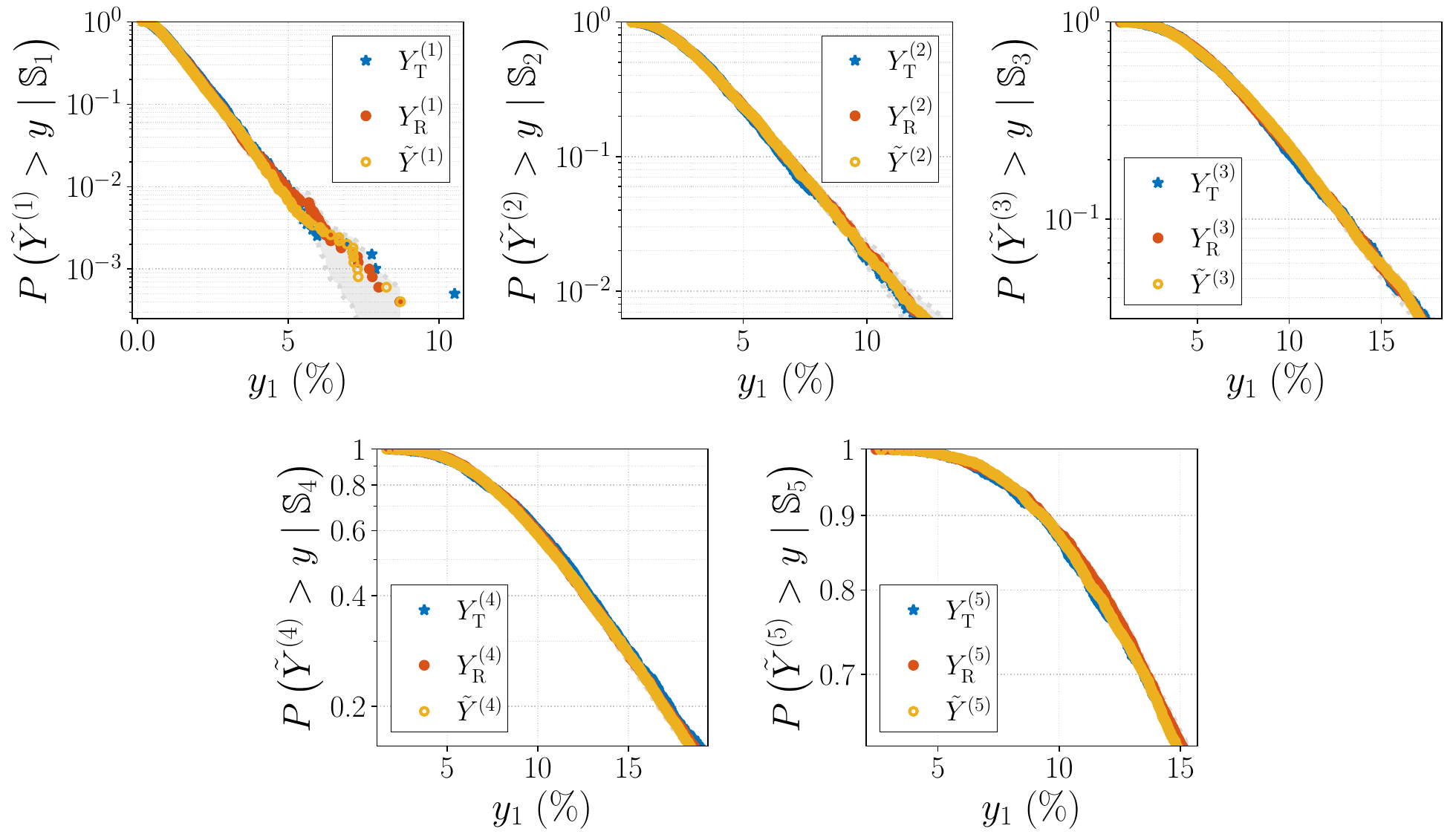}
\caption{Conditional exceedance probabilities for the first-story interstory drift $Y_1$ within each stratum.}
\label{Fig_GSS_SPCE_Drift1_eachStratum}
\end{figure*}

\begin{figure*}[]
\centering
\includegraphics[scale=0.4]{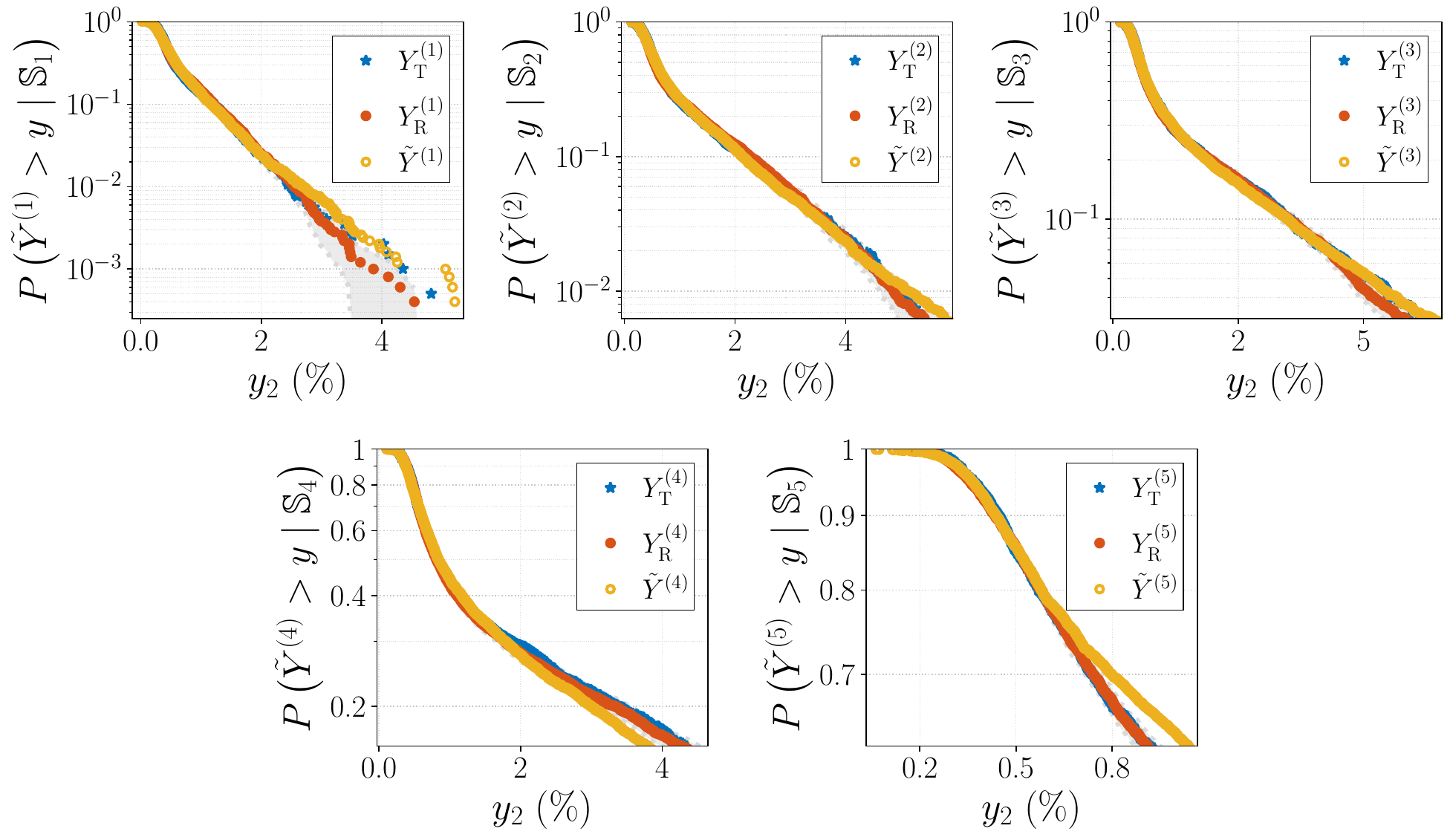}
\caption{Conditional exceedance probabilities for the second-story interstory drift $Y_2$ within each stratum.}
\label{Fig_GSS_SPCE_Drift2_eachStratum}
\end{figure*}

Using the same input vectors $\mathbf{U}^{(i)}$, the trained GSS-SPCEs are evaluated within each stratum to generate the corresponding emulated response samples, denoted by $\widetilde{Y}^{(i)}$. The resulting conditional complementary cumulative distribution functions (CCDFs) are presented in Fig.~\ref{Fig_GSS_SPCE_Drift1_eachStratum} for $Y_1$ and Fig.~\ref{Fig_GSS_SPCE_Drift2_eachStratum} for $Y_2$, together with the reference results $Y_{\mathrm{R}}^{(i)}$ and their associated confidence bounds. The $2000$ structural responses used to train each stratum-specific GSS-SPCE are also shown and denoted by $Y_{\mathrm{T}}^{(i)}$.

The conditional results within each stratum are plotted according to their contribution to the recombined exceedance probability $\widetilde{p}(y)$ defined in Eq.~(\ref{EqPF6}). Because the contribution of the $i$th stratum is weighted by $P(\mathbb{S}_i)$, the conditional exceedance probability $\widetilde{p}^{(i)}(y)$ contributes to the unconditional estimate through the product $\widetilde{p}^{(i)}(y)P(\mathbb{S}_i)$. Therefore, to assess the accuracy of the GSS-SPCE down to a prescribed unconditional probability level $p_{\mathrm{lim}}$, the conditional CCDF within each stratum must be evaluated approximately down to:
\begin{equation}
\widetilde{p}_{\mathrm{lim}}^{(i)}
=
\frac{p_{\mathrm{lim}}}{P(\mathbb{S}_i)}
\label{EqConditionalPlotLimit}
\end{equation}
The plotted range of each conditional CCDF is adjusted according to the probability of occurrence of the corresponding stratum.

For the stratification adopted in this study, with $p_{\mathrm{GSS}}=0.2$ and $n_{\mathbb{S}}=5$, the stratum probabilities are $\{P(\mathbb{S}_i)\}_{i=1}^{5}=\{0.80,0.16,0.032,0.0064,0.0016\}$. Thus, the conditional CCDFs in the tail strata do not need to be represented down to the same probability level as those in the more frequent strata because their smaller stratum weights already reduce their contributions to the unconditional exceedance probability. For instance, adopting $p_{\mathrm{lim}}=10^{-3}$ for the graphical assessment leads to conditional probability limits $\widetilde{p}_{\mathrm{lim}}^{(i)}$ of approximately $1.25\times10^{-3}$, $6.25\times10^{-3}$, $3.13\times10^{-2}$, $1.56\times10^{-1}$, and $6.25\times10^{-1}$ for strata $\mathbb{S}_1$ through $\mathbb{S}_5$, respectively.

The stratum-specific results show good agreement between the reference responses and the samples generated by the GSS-SPCEs over the plotted probability ranges, including the tail regions represented within each stratum. These results support the subsequent use of the weighted recombination to estimate the unconditional exceedance probabilities.

\begin{figure*}[]
\centering
\begin{subfigure}{0.48\textwidth}
\centering
\includegraphics[width=\linewidth]{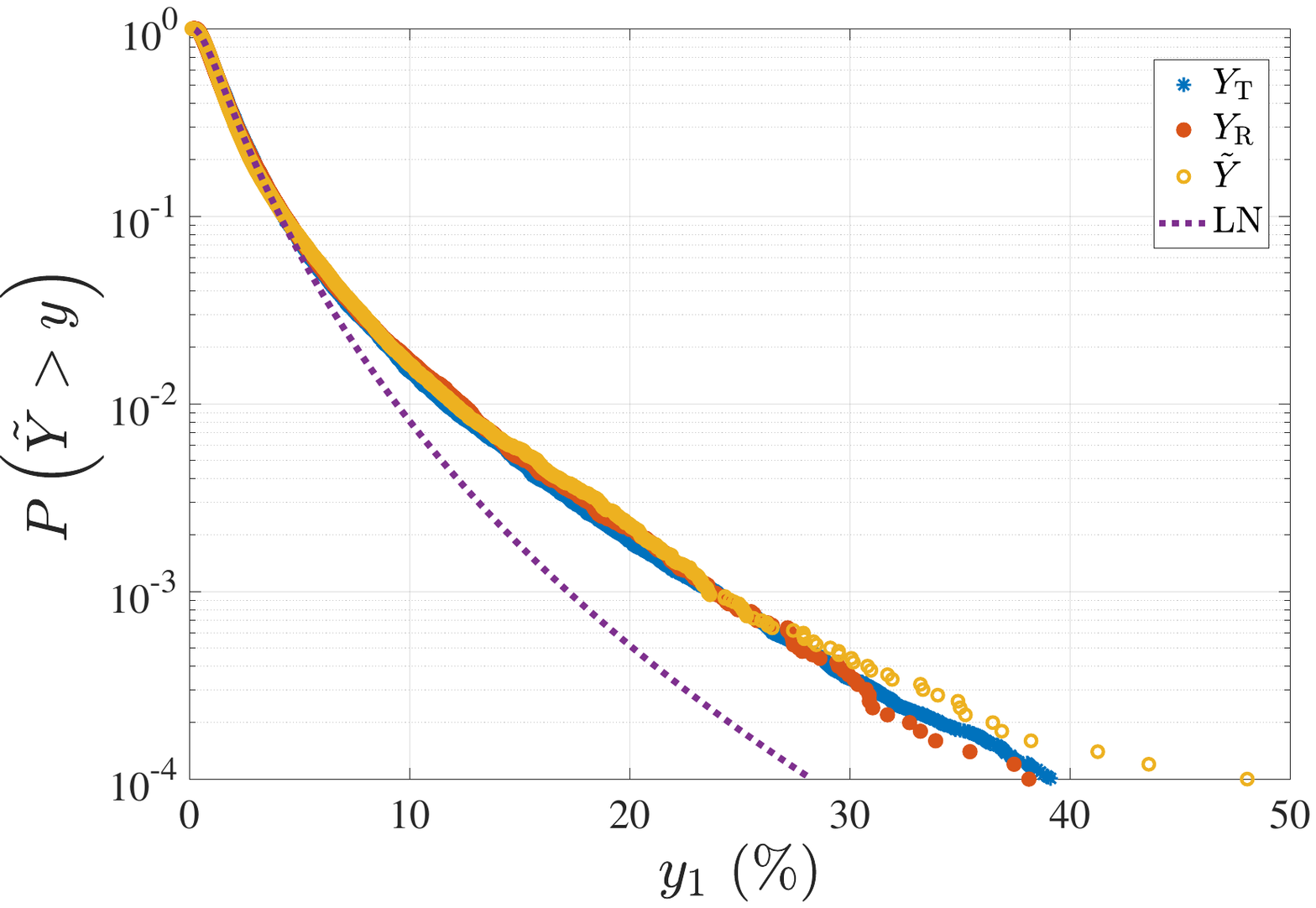}
\caption{First-story interstory drift $Y_1$}
\label{Fig_GSS_SPCE_Drift1}
\end{subfigure}
\hfill
\begin{subfigure}{0.48\textwidth}
\centering
\includegraphics[width=\linewidth]{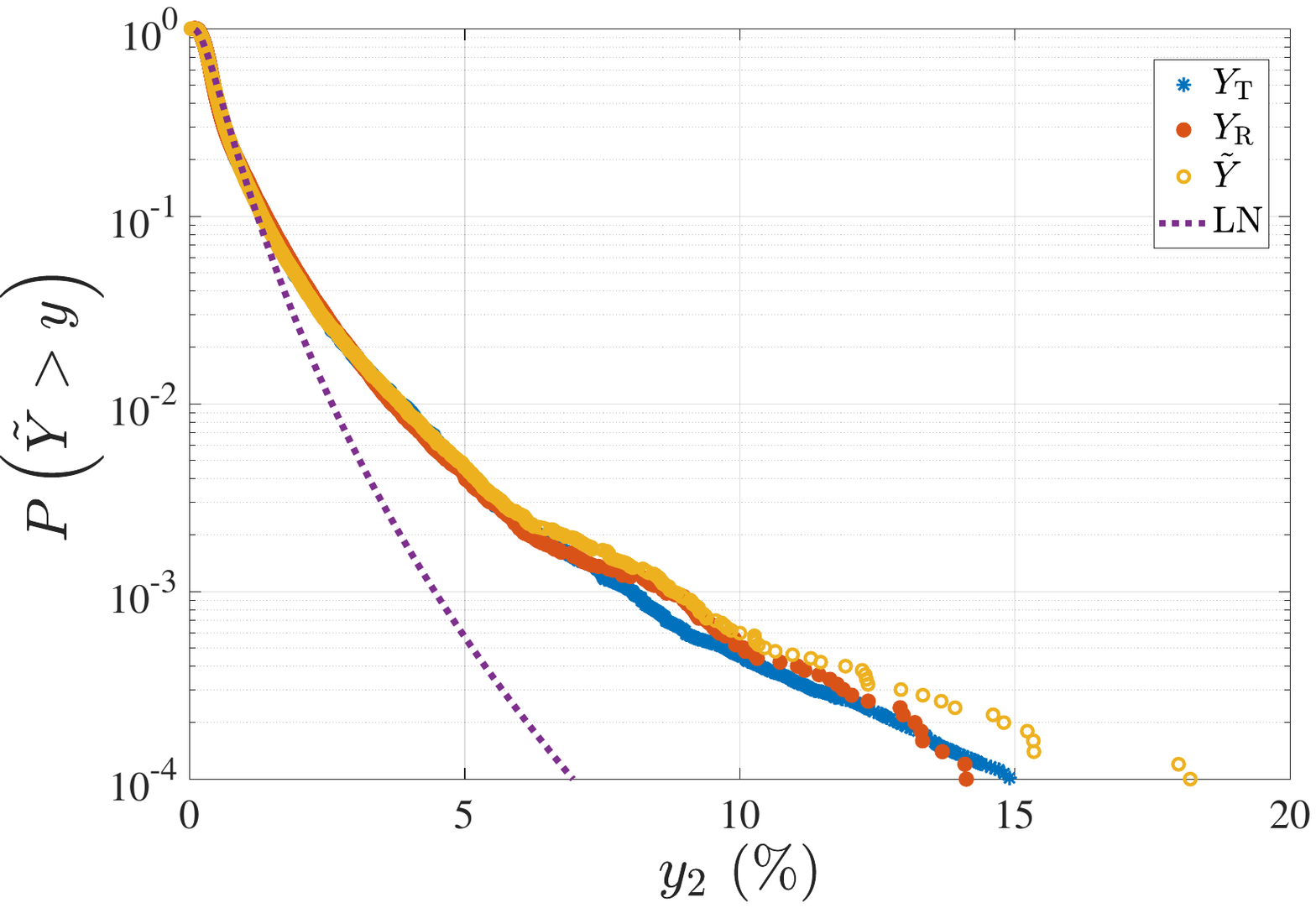}
\caption{Second-story interstory drift $Y_2$}
\label{Fig_GSS_SPCE_Drift2}
\end{subfigure}
\caption{Unconditional exceedance probabilities of the first- and second-story interstory drifts.}
\label{GSS_SPCE_RV}
\end{figure*}

To assess the accuracy of the GSS-SPCE after recombination, an independent reference dataset containing $50{,}000$ NLTHAs is generated using random realizations of the input vector $\mathbf{U}=\left[\mathbf{d}^{T},\mathbf{X}_h^{T}\right]^{T}$. The corresponding reference response samples are denoted by $Y_{\mathrm{R}}$. The same explicit input realizations are then evaluated using the trained stratum-specific SPCEs. For each input realization, the corresponding stratum is identified from the values of $M_w$ and $r$, and the associated SPCE is used to generate one emulated response sample. The resulting emulated response samples are denoted by $\widetilde{Y}$. The empirical unconditional exceedance probability curves are compared in Fig.~\ref{GSS_SPCE_RV}. The figure also includes the weighted recombination of the stratum-specific empirical distributions obtained from the $10{,}000$ NLTHAs used to generate the training responses, denoted by $Y_{\mathrm{T}}$. In addition, a lognormal distribution is fitted to the reference response samples and plotted for each component of $Y$.

The results indicate that the GSS-SPCEs generate structural response samples that accurately represent the reference distributions down to exceedance probabilities of approximately $10^{-3}$. This probability level is consistent with the stratification process because the lower boundary of the final stratum shown in Fig.~\ref{Fig_StrataWiseSamples_a} corresponds to an exceedance rate close to $10^{-3}$. If greater accuracy at lower probabilities is required, additional strata could be introduced to extend the range covered by the stratification. The GSS-SPCE result requires the $10{,}000$ NLTHAs used for training, corresponding to one-fifth of the $50{,}000$ NLTHAs used to construct the reference distribution.

The results also indicate that the lognormal approximation, which is commonly adopted in performance-based engineering applications, begins to deviate from the reference response distributions at exceedance probabilities below approximately $10^{-1}$. This limitation may become more pronounced when the distribution is fitted using fewer response samples because of the computational cost of the nonlinear structural analyses.

To assess the accuracy of GSS-SPCEs trained using fewer support points within each stratum, an error metric focused on the upper tail of the response distribution is defined. Because the reliability assessment is governed by small exceedance probabilities, the discrepancy between the reference and emulated distributions is evaluated only above the $90\%$ quantile of the reference distribution. The dimensionless tail error metric is defined as:
\begin{equation}
\varepsilon_{\mathrm{tail}}
=
\frac{
\mathbb{E}\left[
W_{2,q}^2\left(Y_{\mathrm{R}},\tilde{Y}\right)
\right]
}{
\mathrm{Var}\left[Y_{\mathrm{R}}\right]
},
\qquad q=0.90
\label{EqError}
\end{equation}
where $W_{2,q}^2$ denotes the squared Wasserstein distance restricted to the upper tail of the distributions, defined as:
\begin{equation}
W_{2,q}^2\left(Y_a,Y_b\right)
=
\frac{1}{1-q}
\int_q^1
\left[
Q_a(u)-Q_b(u)
\right]^2
du
\label{EqError2}
\end{equation}
where $Q_a$ and $Q_b$ are the quantile functions of the random variables $Y_a$ and $Y_b$, respectively. The factor $1/(1-q)$ normalizes the metric by the length of the tail interval, so that the error represents the average squared discrepancy between the two quantile functions over the upper tail.

\begin{figure}[]
\centering
\includegraphics[scale=0.29]{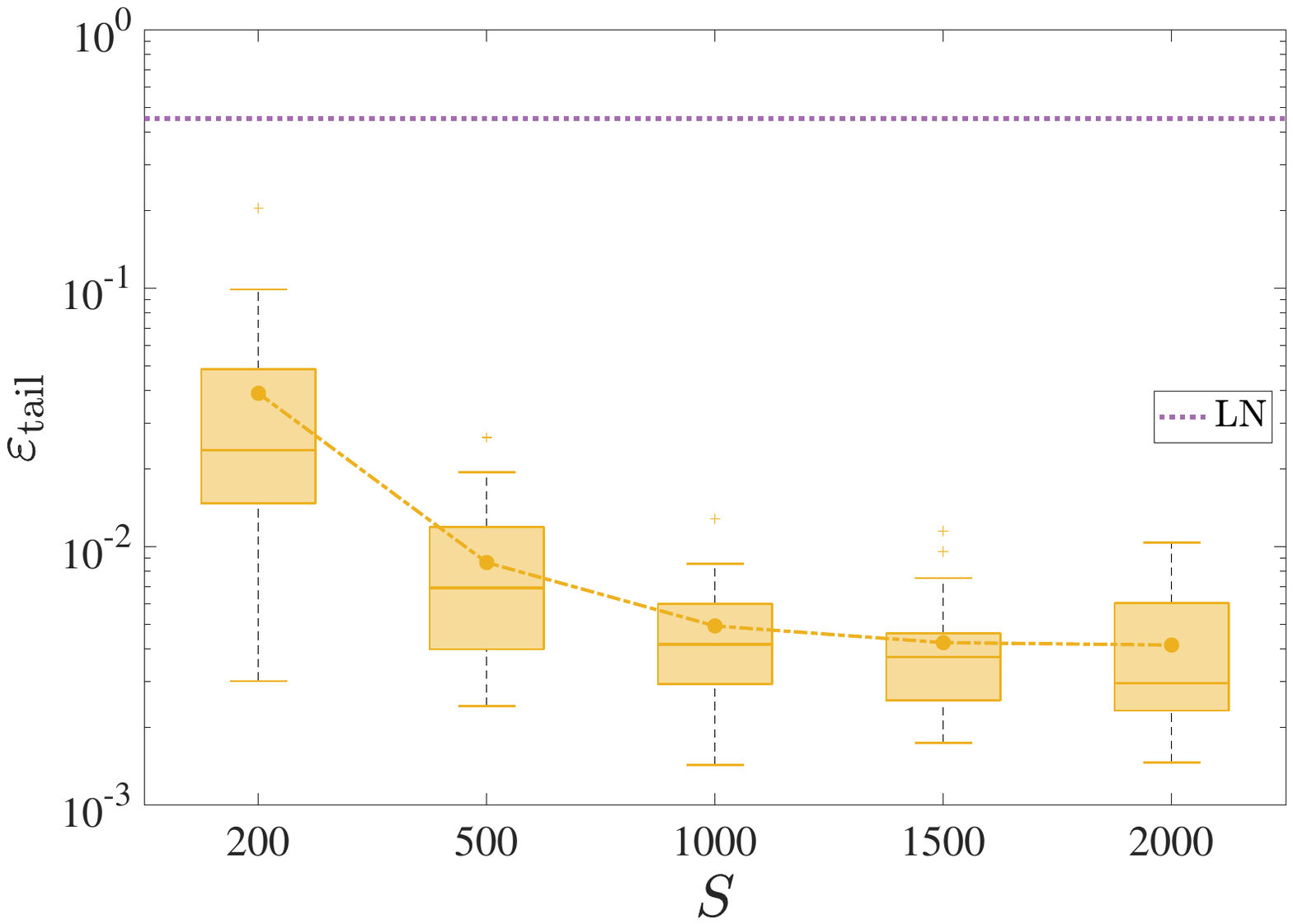}
\caption{Tail error $\varepsilon_{\mathrm{tail}}$ for $Y_1$ as a function of the number of support points $S$ used to train each stratum-specific emulator $\widetilde{Y}^{(i)}$.}
\label{Fig_Error_Drift1}
\end{figure}

For this assessment, each trained GSS-SPCE is evaluated once for each input vector, consistent with the standard use of the emulator after training. The number of support points $S$ within each stratum is varied, and a new set of stratum-specific SPCEs is trained for each value of $S$. To account for variability arising from the selection of the DoE and the intrinsic stochasticity of the simulator, each scenario is repeated 20 times using independent DoEs. The resulting values of $\varepsilon_{\mathrm{tail}}$ are summarized using box plots in Fig.~\ref{Fig_Error_Drift1} for the first-story interstory drift $Y_1$. The dash-dotted line connects the mean errors obtained from the 20 repetitions for each value of $S$, while the horizontal line represents the error obtained by approximating the reference response distribution as lognormal.

The results indicate that using $S=1500$, or even $S=1000$, does not materially degrade the tail accuracy relative to that obtained using $S=2000$. This suggests that the number of support points used to train each stratum-specific emulator could be reduced by as much as one-half while maintaining an adequate level of accuracy. Such a reduction would further improve the computational efficiency of the proposed framework.

% ----------------------

\subsubsection{Optimization results}
\label{subsubsec:gssspceresultsOt}

The results obtained by solving the optimization problem defined in Eq.~(\ref{EqCaseStudy}) are presented in this section. For each design candidate $\mathbf{d}^{(\ell)}$, the probabilistic constraints associated with $\widetilde{p}_1\left(y_{\mathrm{cr}}\mid\mathbf{d}^{(\ell)}\right)$ and $\widetilde{p}_2\left(y_{\mathrm{cr}}\mid\mathbf{d}^{(\ell)}\right)$ are evaluated using the trained stratum-specific SPCEs. This process is repeated until the termination criteria of the optimization algorithm are satisfied and the optimal design $\mathbf{d}^{\star}$ is obtained.

The optimization problem is solved using a genetic algorithm (GA) \cite{holland1992adaptation}, which explores the design space through selection, crossover, and mutation operations. Although a gradient-based algorithm could also be considered, the probabilistic constraints are implicit functions of the design variables, as discussed in Section~\ref{subsec:otproblem}, and their gradients are not readily available. The GA is therefore adopted because it does not require constraint gradients and is suitable for optimization problems with potentially nonconvex constraint functions.

A population size of 40 candidates is used. For each generation, $50{,}000$ random realizations of the hazard-related variables $\mathbf{X}_h$ are generated and combined with each design candidate $\mathbf{d}^{(\ell)}$ to form the corresponding input vectors $\mathbf{U}=\left[\left(\mathbf{d}^{(\ell)}\right)^T,\mathbf{X}_h^T\right]^T$. For each realization, the corresponding stratum $\mathbb{S}_i$ is identified from the values of $M_w$ and $r$, and the associated stratum-specific SPCEs are evaluated to generate realizations of the emulated responses $\widetilde{Y}_j^{(i)}$, for $j=1,2$. The conditional exceedance probabilities $\widetilde{p}_j^{(i)}\left(y_{\mathrm{cr}}\mid\mathbf{d}^{(\ell)}\right)$ are then estimated within each stratum and recombined according to Eq.~(\ref{EqPF6}) to obtain the unconditional exceedance probabilities $\widetilde{p}_j\left(y_{\mathrm{cr}}\mid\mathbf{d}^{(\ell)}\right)$.

\begin{figure*}[]
\centering
\begin{subfigure}{0.48\textwidth}
\centering
\includegraphics[width=\linewidth]{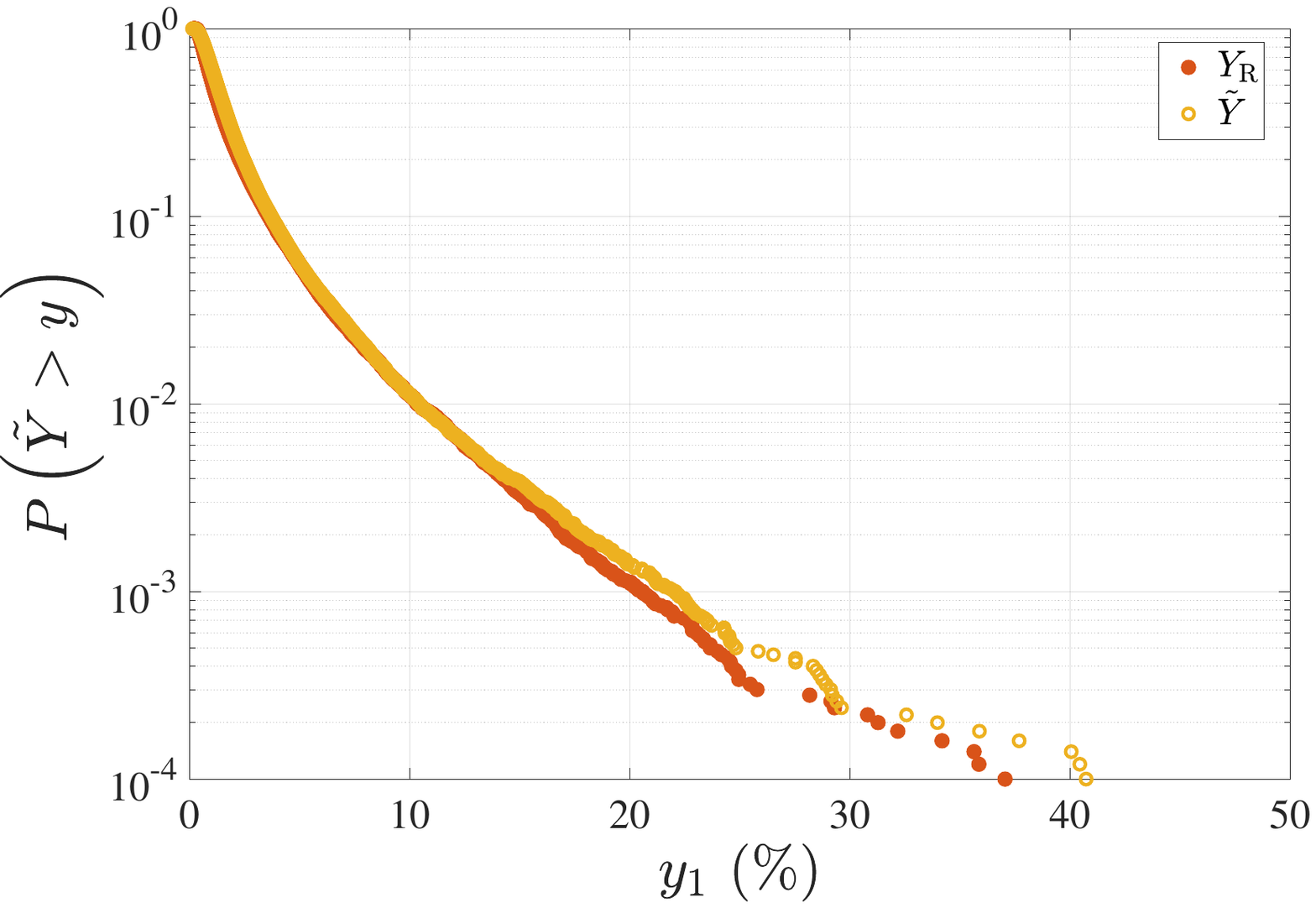}
\caption{First-story interstory drift $Y_1$}
\label{Fig_GSS_SPCE_Ot_Drift1}
\end{subfigure}
\hfill
\begin{subfigure}{0.48\textwidth}
\centering
\includegraphics[width=\linewidth]{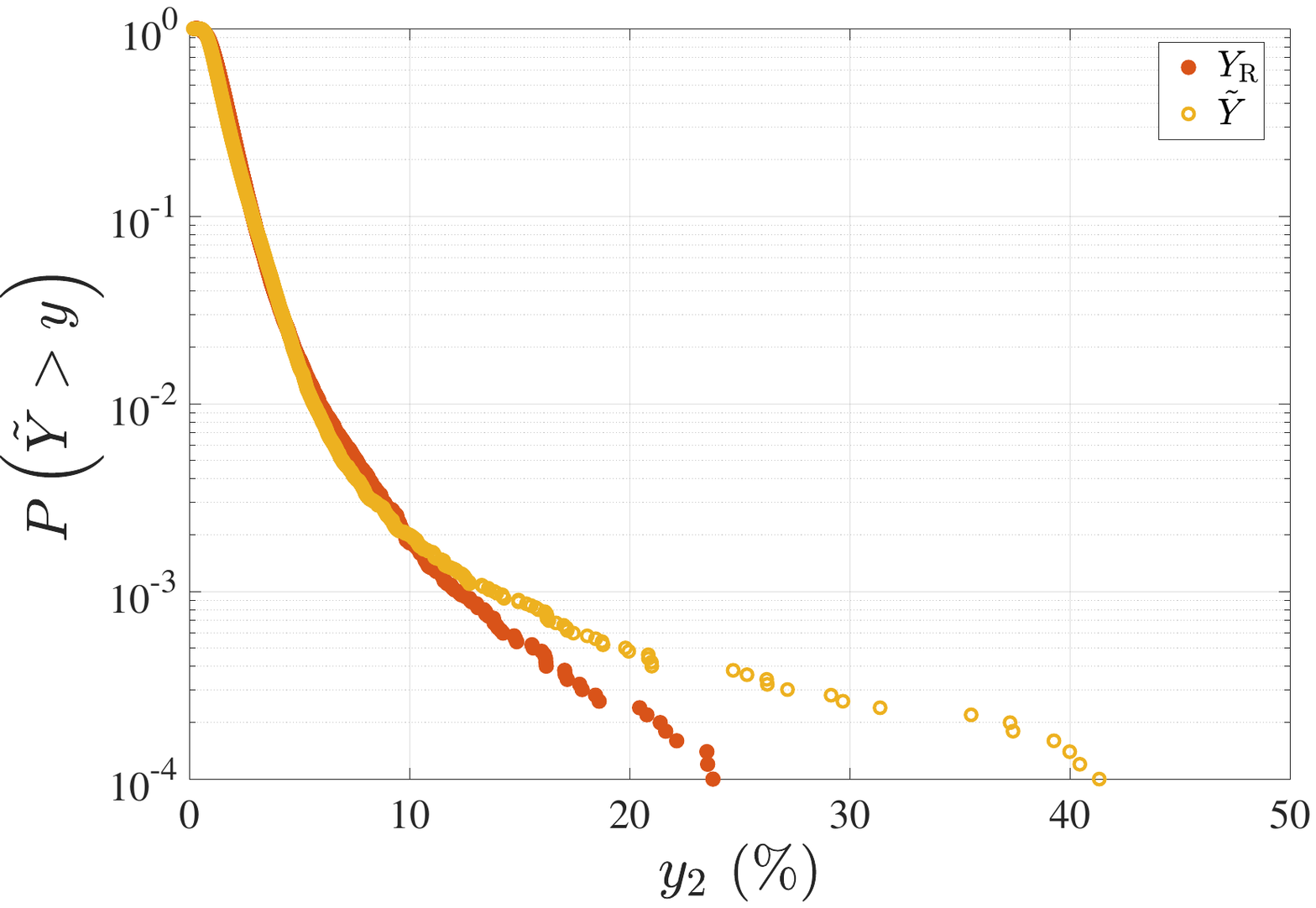}
\caption{Second-story interstory drift $Y_2$}
\label{Fig_GSS_SPCE_Ot_Drift2}
\end{subfigure}
\caption{Unconditional exceedance probabilities of the first- and second-story interstory drifts for $\mathbf{d}^{\star}=\left[39.14,20\right]^T\,\mathrm{cm}^2$.}
\label{GSS_SPCE_Otresults}
\end{figure*}

The optimal design obtained is $\mathbf{d}^{\star}=\left[39.14,20\right]^T\,\mathrm{cm}^2$. For this design, an independent reference dataset containing $50{,}000$ response realizations is generated through NLTHA using random realizations of the hazard-related variables $M_w$ and $r$. The resulting reference responses are denoted by $Y_{\mathrm{R}}$. The same $50{,}000$ explicit input realizations are then evaluated using the trained stratum-specific SPCEs to generate the corresponding emulated responses, denoted by $\widetilde{Y}$. The resulting unconditional exceedance probability curves are presented in Fig.~\ref{GSS_SPCE_Otresults} for the first- and second-story interstory drifts, shown in Figs.~\ref{Fig_GSS_SPCE_Ot_Drift1} and~\ref{Fig_GSS_SPCE_Ot_Drift2}, respectively.
The results show good agreement between the GSS-SPCE estimates and the reference distributions for the optimal design $\mathbf{d}^{\star}$ down to exceedance probabilities of approximately $10^{-3}$. This probability level is below the acceptable exceedance probability adopted in the optimization problem, $p_{\mathrm{t}}=4\times10^{-3}$, indicating that the probabilistic constraints are accurately evaluated over the probability range of interest. These results support the use of the GSS-SPCE approach for optimization problems involving acceptable exceedance probabilities within the range validated in this study, which would otherwise require a large number of evaluations of the nonlinear structural model.

Regarding computational effort, the GA terminated after 870 evaluations of the objective function and probabilistic constraints. The termination criterion was satisfied when the average change in the fitness value fell below the prescribed function tolerance while the maximum constraint violation remained below the constraint tolerance of $10^{-5}$. Within the proposed framework, the probabilistic constraints are evaluated using the trained GSS-SPCEs. Therefore, no additional evaluations of the nonlinear structural model $g_{\mathrm{NL}}$ are required during the optimization loop.

To provide an estimate of the computational savings achieved by the proposed framework, two hypothetical direct optimization strategies are considered. For this comparison, the same 870 design-candidate evaluations are assumed for all strategies. This assumption enables a consistent comparison of the required numbers of nonlinear model evaluations, although the convergence behavior of the optimization algorithm could differ when the probabilistic constraints are estimated directly through simulation. Without the emulators, each design candidate would require the direct propagation of samples through $g_{\mathrm{NL}}$ to estimate the exceedance probabilities associated with the probabilistic constraints. If a direct MC-based optimization were performed using $50{,}000$ samples per design candidate, the optimization would require approximately $870\times50{,}000=4.35\times10^7$ evaluations of $g_{\mathrm{NL}}$. A direct GSS-based optimization would reduce this number by estimating the exceedance probabilities through stratified sampling. Ideally, the number of samples assigned to each stratum would be determined using an optimal allocation scheme, as proposed by \citet{arunachalam2023generalized}. Such an allocation depends on the conditional exceedance probabilities within the individual strata, which are not known a priori and may vary with the design candidate. Therefore, for the purpose of this illustrative comparison, 100 nonlinear model evaluations are assumed within each of the five strata. This results in $5\times100=500$ evaluations of $g_{\mathrm{NL}}$ per design candidate and approximately $870\times500=4.35\times10^5$ evaluations over the optimization process. %This simplified allocation is used only to estimate the order of magnitude of the computational effort and does not necessarily provide the same statistical accuracy as the other approaches.
In contrast, constructing the GSS-SPCEs requires a one-time computational cost of $S=2000$ support points within each of the $n_{\mathbb{S}}=5$ strata, resulting in $5\times2000=10{,}000$ evaluations of $g_{\mathrm{NL}}$. No additional nonlinear model evaluations are required during the subsequent optimization. Under the assumptions adopted for this comparison, the GSS-SPCE framework reduces the number of nonlinear model evaluations by an estimated factor of $4.35\times10^7/10{,}000=4350$ relative to direct MC-based optimization and by an estimated factor of $4.35\times10^5/10{,}000=43.5$ relative to direct GSS-based optimization. The corresponding computational requirements are summarized in Table~\ref{Tab_Comp_Efficiency}. %The nonlinear analyses performed solely for validation of the proposed framework are not included in these counts.
\begin{table*}[]
\centering
\scriptsize
%\footnotesize
\caption{Estimated numbers of nonlinear model evaluations required by the different optimization strategies.}
\label{Tab_Comp_Efficiency}
\begin{tabular}{lccccc}
\hline
\shortstack{Optimization\\approach}
&
\shortstack{Evaluations per\\design candidate}
&
\shortstack{Design-candidate\\evaluations}
&
\shortstack{One-time training\\evaluations}
&
\shortstack{Total evaluations\\of $g_{\mathrm{NL}}$}
&
\shortstack{Total relative\\to GSS-SPCE}
\\
\hline
Direct MC
& $50{,}000$
& $870$
& $0$
& $4.35\times10^{7}$
& $4350$
\\
GSS
& $5\times100=500$
& $870$
& $0$
& $4.35\times10^{5}$
& $43.5$
\\
GSS-SPCE
& $0$
& $870$
& $1.00\times10^{4}$
& $1.00\times10^{4}$
& $1.0$
\\
\hline
\end{tabular}
\end{table*}

It is also informative to compare the optimization results obtained in this study with the original design reported in NIST GCR 10-917-8 \cite{nist2010evaluation}. Because the design objectives and performance criteria adopted in the two procedures differ, this comparison is intended only to provide a reference point for interpreting the optimized design. As shown in Table~\ref{Tab_NIST_Comparison}, the optimized design has smaller BRB areas than the original NIST design while satisfying the probabilistic constraints within the tolerance adopted by the optimization algorithm. The optimized design also has a lower estimated initial construction cost under the cost model adopted in this study. These results indicate that the proposed PBRO formulation can identify a more economical design while maintaining the prescribed probabilistic seismic performance requirements considered in this study. 

\begin{table*}[]
\centering
\footnotesize
\caption{Comparison of the original NIST design and the optimized design obtained from the GSS-SPCE framework. Here, $\widetilde{y}_{j,p_{\mathrm{t}}}$ is the interstory drift corresponding to an exceedance probability of $p_{\mathrm{t}}$ for story $j$.}
\label{Tab_NIST_Comparison}
\begin{tabular}{lccccccc}
\hline
Design
& $d_1$ ($\mathrm{cm}^2$)
& $d_2$ ($\mathrm{cm}^2$)
& $\widetilde{p}_1(y_{\mathrm{cr}})$
& $\widetilde{p}_2(y_{\mathrm{cr}})$
& $\widetilde{y}_{1,p_{\mathrm{t}}}$ ($\%$)
& $\widetilde{y}_{2,p_{\mathrm{t}}}$ ($\%$)
& $C\left(\mathbf{d}\right)$ (\$)
\\
\hline
NIST design
& 58.06
& 38.71
& $2.67\times10^{-3}$
& $5.40\times10^{-5}$
& 13.22
& 3.14
& $2.16\times10^{4}$
\\
Optimized design
& 39.14
& 20.00
& $3.84\times10^{-3}$
& $8.76\times10^{-4}$
& 14.69
& 7.56
& $1.94\times10^{4}$
\\
\hline
\end{tabular}
\end{table*}
%

% ----------------
% ----------------
% ----------------

\section{Conclusions}
\label{sec:conclusions}

This paper proposes a framework that combines an advanced stratified sampling scheme with stochastic emulation to estimate exceedance probabilities of structural responses under stochastic excitation, with particular emphasis on accurately representing distribution tails. The proposed GSS-SPCE framework is intended for use within PBRO problems in which the constraints are defined in terms of acceptable exceedance probabilities. GSS partitions the input space according to hazard intensity, while independent SPCEs are trained within the resulting strata to emulate the corresponding conditional response distributions. The stratum-specific conditional exceedance probabilities are then recombined using the law of total probability, providing an efficient means of evaluating the probabilistic constraints throughout the optimization process.

The proposed framework is evaluated through a case study involving the optimization of the BRB areas in a two-story BRBF. The objective is to minimize the estimated initial construction cost while satisfying prescribed probabilistic constraints on the first- and second-story interstory drifts under seismic excitation. The results show that the GSS-SPCE accurately represents the response distributions over the probability range relevant to the optimization, including their upper tails. The optimized design satisfies the prescribed probabilistic constraints while requiring smaller BRB areas and having a lower estimated initial construction cost than the original reference design. Under the assumptions adopted for the computational comparison, the proposed framework requires approximately 4000 times fewer nonlinear model evaluations than direct Monte Carlo-based optimization and 40 times fewer than direct GSS-based optimization.

The proposed framework provides a computationally efficient approach for optimizing structures under stochastic excitation when response distributions and exceedance probabilities must be evaluated repeatedly for different design candidates. Its use may also be extended to broader performance-based engineering applications involving probabilistic performance assessment under uncertainty. Future developments may include applications to higher-dimensional design problems, the consideration of multiple performance objectives and constraints, the use of gradient-based optimization methods, and the assessment of alternative stratification variables for different natural hazards.

\section*{Nomenclature}

\begin{itemize}[
    leftmargin=3.2cm,
    labelwidth=2.7cm,
    labelsep=0.5cm,
    align=left,
    itemsep=0pt,
    parsep=0pt,
    topsep=2pt,
    partopsep=0pt
]

    % General probability and response quantities
    \item[$Y$] Structural response
    \item[$\widetilde{Y}$] Emulated structural response
    \item[$Y_{\mathrm{R}}$] Reference structural response
    \item[$Y_{\mathrm{T}}$] Training structural response
    \item[$y$] Response level
    \item[$y_{\mathrm{cr}}$] Critical response threshold
    \item[$p(y)$] Exceedance probability associated with response level $y$
    \item[$\widetilde{p}(y)$] Emulated exceedance probability
    \item[$p_{\mathrm{t}}$] Acceptable exceedance probability
    \item[$p_{\mathrm{lim}}$] Prescribed unconditional probability level
    \item[$F$] Cumulative distribution function
    \item[$f$] Probability density function

    % Design and cost quantities
    \item[$\mathbf{d}$] Vector of design variables
    \item[$\mathcal{D}$] Design space
    \item[$C(\mathbf{d})$] Construction cost
    \item[$W_{\mathrm{b}}$] Total weight of the beams
    \item[$W_{\mathrm{c}}$] Total weight of the columns
    \item[$W_{\mathrm{BRB}}$] Total weight of the BRBs

    % Random variables and uncertainty representation
    \item[$\mathbf{X}$] Vector of explicit random input variables
    \item[$\mathcal{D}_{\mathbf{X}}$] Input space of $\mathbf{X}$
    \item[$\mathbf{X}_m$] Vector of model-related random variables
    \item[$\mathbf{X}_h$] Vector of hazard-related random variables
    \item[$\mathbf{X}_w$] Vector representing the stochastic variability of the load excitation
    \item[$\boldsymbol{\Xi}$] Complete vector of uncertainties involved in the structural performance assessment
    \item[$n_{\mathrm{RV}}$] Number of random variables
    \item[$M_w$] Earthquake moment magnitude
    \item[$r$] Epicentral distance

    % Stochastic and deterministic models
    \item[$g_{\mathrm{NL}}$] Nonlinear structural response model
    \item[$\mathcal{M}_s$] Stochastic simulator
    \item[$\mathcal{M}_d$] Deterministic simulator
    \item[$\mathcal{H}$] Auxiliary model used to evaluate the stratification variable
    \item[$\Omega$] Sample space associated with the internal stochasticity of the simulator
    \item[$\omega$] Event in $\Omega$

    % SPCE and DoE quantities
    \item[$\mathbf{U}$] Augmented explicit input vector composed of design and random variables
    \item[$Z$] Latent random variable representing intrinsic simulator variability
    \item[$\epsilon$] Additive SPCE noise term
    \item[$\sigma$] Standard deviation of the SPCE noise term
    \item[$\boldsymbol{\alpha}$] Polynomial multi-index
    \item[$c_{\boldsymbol{\alpha}}$] SPCE expansion coefficient
    \item[$\psi_{\boldsymbol{\alpha}}$] Multivariate polynomial basis function
    \item[$\mathcal{A}$] Finite set of polynomial multi-indices
    \item[$S$] Number of support points within each stratum
    \item[$\mathcal{X}_S$] DoE matrix of explicit random inputs
    \item[$\mathcal{Y}_S$] Simulator evaluations associated with $\mathcal{X}_S$
    \item[$\mathcal{U}_S^{(i)}$] Stratum-specific DoE in the augmented input space

    % GSS quantities
    \item[$\boldsymbol{\tau}$] Subset of hazard-related variables used for stratification
    \item[$\mathcal{V}$] Stratification variable
    \item[$\mathcal{V}_i$] Boundary of the $i$th stratum
    \item[$\mathbb{S}_i$] $i$th stratum
    \item[$P(\mathbb{S}_i)$] Probability of occurrence of the $i$th stratum
    \item[$n_{\mathbb{S}}$] Number of strata
    \item[$p_{\mathrm{GSS}}$] Probability constant used in the GSS scheme
    \item[$\hat{n}$] Number of samples used to determine the stratum boundaries

    % Error metric quantities
    \item[$\varepsilon_{\mathrm{tail}}$] Dimensionless upper-tail error metric
    \item[$W_{2,q}^{2}$] Squared Wasserstein distance restricted to the upper tail
    \item[$q$] Quantile defining the lower boundary of the upper tail
    \item[$Q_a,Q_b$] Quantile functions of the random variables $Y_a$ and $Y_b$

\end{itemize}

% ---------------------
% ---------------------

\section*{Declaration of competing interest}
\label{sec:Declaration of competing interest}

The authors have no conflict of interest in submitting this manuscript.

\section*{Data availability}
\label{sec:Data availability}

Data will be made available on request.

\section*{Acknowledgment}
\label{sec:Acknowledgment}

This research was supported in part by the Brazilian agencies FAPESP (São Paulo Research Foundation, grants 2020/14072-7, 2023/02504-8, and 2025/13211-7), CAPES (Coordenação de Aperfeiçoamento de Pessoal de Nível Superior, Finance Code 001), and CNPq (National Council for Scientific and Technological Development, grant No. 309107/2020-2). The contribution of the second author is based upon work supported in part by the U.S. National Science Foundation under Award No. CMMI-2118488. Any opinions, findings, and conclusions or recommendations expressed in this material are those of the authors and do not necessarily reflect the views of the U.S. National Science Foundation.

%\appendix

%\section{Sample Appendix Section}

\appendix

\section{Seismic hazard and stochastic ground-motion model}
\label{sec:sample:SGMM}

The stochastic ground-motion model used to generate synthetic records that are statistically consistent with the target spectrum employs a two-corner point-source formulation that captures the physics of fault rupture and wave propagation \cite{vetter2012global,arunachalam2023generalized,atkinson2000stochastic}. Nonstationarity in time is incorporated through a temporal envelope function, while the frequency content is controlled by a parametric radiation spectrum $A(f;M_w,r)$, where $f$ denotes frequency. Both depend on the earthquake magnitude $M_w$ and epicentral distance $r$.

The frequency-domain radiation spectrum is expressed as:
\begin{equation}
A(f;M_w,r)
=
\left(2\pi f\right)^2
E(f;M_w)
P(f;r)
G(f)
\label{EqSGMM1}
\end{equation}
where $E(f;M_w)$ represents the source term, $P(f;r)$ models path effects, and $G(f)$ accounts for site amplification.

The temporal envelope function, which governs the evolution of the ground motion over time, is given by \cite{boore2003simulation}:
\begin{equation}
e(t;M_w,r)
=
a_t
\left(
\frac{t}{t_n}
\right)^{b_t}
\exp\left(
-c_t\frac{t}{t_n}
\right)
\label{EqSGMM2}
\end{equation}
where $t$ denotes time, $t_n$ is the ground-motion duration, and the constants $a_t$, $b_t$, and $c_t$ are selected such that $e(t;M_w,r)$ peaks at $t=\lambda_t t_n$ with a normalized maximum of 1 and satisfies $e(t_n;M_w,r)=\eta_t$.

Synthetic ground motions are generated by modulating the white-noise sequence $\mathbf{X}_w=[X_w(i\Delta t)]$, where $\Delta t$ is the time increment, as follows:

\begin{enumerate}[noitemsep,nolistsep]
    \item Multiply $\mathbf{X}_w$ by the envelope function $e(t;M_w,r)$.
    \item Transform the resulting sequence to the frequency domain.
    \item Normalize by the square root of the mean-square amplitude spectrum.
    \item Multiply by the radiation spectrum $A(f;M_w,r)$.
    \item Transform back to the time domain to obtain the acceleration time history.
\end{enumerate}

The model parameters used for ground-motion generation follow \cite{vetter2012global,arunachalam2023generalized}: radiation pattern $R_\Phi=0.55$, shear-wave velocity $\bar{V}_s=3.5\,\mathrm{km/s}$, rock density $\rho_s=2.8\,\mathrm{g/cm^3}$, seismic velocity $c_Q=3.5\,\mathrm{km/s}$, and anelastic attenuation function $Q(f)=180f^{0.45}$. Geometric spreading is modeled as $Z(R)=1/R$ for $R<70\,\mathrm{km}$ and $Z(R)=1/70$ for $R\geq70\,\mathrm{km}$, where $R$ is the source-to-site distance. Site amplification follows the generic rock-site model of \citet{boore1997site}. The envelope-function parameters are set to $\lambda_t=0.2$ and $\eta_t=0.05$ \cite{boore2003simulation}. The high-dimensional vector $\mathbf{X}_w$ represents RTR variability, while $M_w$ and $r$ constitute the hazard-related random vector $\mathbf{X}_h$ and represent the dominant seismic hazard parameters \cite{vetter2012global,arunachalam2023generalized}.

% \clearpage
\bibliographystyle{elsarticle-num-names} 
\bibliography{reference}

\begin{thebibliography}{65}
\expandafter\ifx\csname natexlab\endcsname\relax\def\natexlab#1{#1}\fi
\providecommand{\url}[1]{\texttt{#1}}
\providecommand{\href}[2]{#2}
\providecommand{\path}[1]{#1}
\providecommand{\DOIprefix}{doi:}
\providecommand{\ArXivprefix}{arXiv:}
\providecommand{\URLprefix}{URL: }
\providecommand{\Pubmedprefix}{pmid:}
\providecommand{\doi}[1]{\href{http://dx.doi.org/#1}{\path{#1}}}
\providecommand{\Pubmed}[1]{\href{pmid:#1}{\path{#1}}}
\providecommand{\bibinfo}[2]{#2}
\ifx\xfnm\relax \def\xfnm[#1]{\unskip,\space#1}\fi
%Type = Inproceedings
\bibitem[{Krawinkler(1999)}]{krawinkler1999challenges}
\bibinfo{author}{H.~Krawinkler},
\newblock \bibinfo{title}{Challenges and progress in performance-based earthquake engineering},
\newblock in: \bibinfo{booktitle}{International Seminar on Seismic Engineering for Tomorrow--In Honor of Professor Hiroshi Akiyama}, volume~\bibinfo{volume}{26}, \bibinfo{year}{1999}.
%Type = Article
\bibitem[{Spence and Arunachalam(2022)}]{spence2022performance}
\bibinfo{author}{S.~M. Spence}, \bibinfo{author}{S.~Arunachalam},
\newblock \bibinfo{title}{Performance-based wind engineering: Background and state of the art},
\newblock \bibinfo{journal}{Frontiers in Built Environment} \bibinfo{volume}{8} (\bibinfo{year}{2022}) \bibinfo{pages}{830207}.
%Type = Article
\bibitem[{Hassanzadeh et~al.(024a)Hassanzadeh, Moradi, and Burton}]{hassanzadeh2024performance}
\bibinfo{author}{A.~Hassanzadeh}, \bibinfo{author}{S.~Moradi}, \bibinfo{author}{H.~V. Burton},
\newblock \bibinfo{title}{Performance-based design optimization of structures: State-of-the-art review},
\newblock \bibinfo{journal}{Journal of Structural Engineering} \bibinfo{volume}{150} (\bibinfo{year}{2024a}) \bibinfo{pages}{03124001}.
%Type = Article
\bibitem[{Rodrigues et~al.(2026)Rodrigues, Spence, and Beck}]{rodrigues2026incorporating}
\bibinfo{author}{I.~D. Rodrigues}, \bibinfo{author}{S.~M. Spence}, \bibinfo{author}{A.~T. Beck},
\newblock \bibinfo{title}{Incorporating statistical uncertainty into metamodel-driven performance-based risk optimization for seismic design},
\newblock \bibinfo{journal}{Engineering Structures} \bibinfo{volume}{348} (\bibinfo{year}{2026}) \bibinfo{pages}{121744}.
%Type = Article
\bibitem[{Beck and de~Santana~Gomes(2012)}]{beck2012comparison}
\bibinfo{author}{A.~T. Beck}, \bibinfo{author}{W.~J. de~Santana~Gomes},
\newblock \bibinfo{title}{A comparison of deterministic, reliability-based and risk-based structural optimization under uncertainty},
\newblock \bibinfo{journal}{Probabilistic Engineering Mechanics} \bibinfo{volume}{28} (\bibinfo{year}{2012}) \bibinfo{pages}{18--29}.
%Type = Article
\bibitem[{Spence and Kareem(2014)}]{spence2014performance}
\bibinfo{author}{S.~M. Spence}, \bibinfo{author}{A.~Kareem},
\newblock \bibinfo{title}{Performance-based design and optimization of uncertain wind-excited dynamic building systems},
\newblock \bibinfo{journal}{Engineering Structures} \bibinfo{volume}{78} (\bibinfo{year}{2014}) \bibinfo{pages}{133--144}.
%Type = Article
\bibitem[{Rastegaran et~al.(2022)Rastegaran, Aval, and Sangalaki}]{rastegaran2022multi}
\bibinfo{author}{M.~Rastegaran}, \bibinfo{author}{S.~B. Aval}, \bibinfo{author}{E.~Sangalaki},
\newblock \bibinfo{title}{Multi-objective reliability-based seismic performance design optimization of {SMRF}s considering various sources of uncertainty},
\newblock \bibinfo{journal}{Engineering Structures} \bibinfo{volume}{261} (\bibinfo{year}{2022}) \bibinfo{pages}{114219}.
%Type = Article
\bibitem[{Ni et~al.(2024)Ni, Yuan, Fu, Bai, and Liu}]{NI2024103707}
\bibinfo{author}{P.~Ni}, \bibinfo{author}{Z.~Yuan}, \bibinfo{author}{J.~Fu}, \bibinfo{author}{Y.~Bai}, \bibinfo{author}{L.~Liu},
\newblock \bibinfo{title}{Stochastic design optimization of nonlinear structures under random seismic excitations using incremental dynamic analysis},
\newblock \bibinfo{journal}{Probabilistic Engineering Mechanics} \bibinfo{volume}{78} (\bibinfo{year}{2024}) \bibinfo{pages}{103707}.
%Type = Article
\bibitem[{Lopez et~al.(2025)Lopez, dos Santos, and Miguel}]{lopez2025efficient}
\bibinfo{author}{R.~H. Lopez}, \bibinfo{author}{K.~R. dos Santos}, \bibinfo{author}{L.~F.~F. Miguel},
\newblock \bibinfo{title}{An efficient approach for taking into account uncertainties in structural parameters in performance-based design optimization},
\newblock \bibinfo{journal}{Engineering Structures} \bibinfo{volume}{336} (\bibinfo{year}{2025}) \bibinfo{pages}{120382}.
%Type = Article
\bibitem[{Movaghar et~al.(2025)Movaghar, Taflanidis, Giaralis, and Vamvatsikos}]{movaghar2025}
\bibinfo{author}{P.~T. Movaghar}, \bibinfo{author}{A.~A. Taflanidis}, \bibinfo{author}{A.~Giaralis}, \bibinfo{author}{D.~Vamvatsikos},
\newblock \bibinfo{title}{Sustainability-driven risk-based design of inerter vibration absorbers for multistory hysteretic buildings under seismic hazard},
\newblock \bibinfo{journal}{Journal of Engineering Mechanics} \bibinfo{volume}{151} (\bibinfo{year}{2025}) \bibinfo{pages}{04025058}.
%Type = Article
\bibitem[{Spence(2018)}]{spence2018optimization}
\bibinfo{author}{S.~M.~J. Spence},
\newblock \bibinfo{title}{Optimization of uncertain and dynamic high-rise structures for occupant comfort: An adaptive kriging approach},
\newblock \bibinfo{journal}{Structural Safety} \bibinfo{volume}{75} (\bibinfo{year}{2018}) \bibinfo{pages}{57--66}.
%Type = Article
\bibitem[{Kim and Song(2021)}]{kim2021reliability}
\bibinfo{author}{J.~Kim}, \bibinfo{author}{J.~Song},
\newblock \bibinfo{title}{Reliability-based design optimization using quantile surrogates by adaptive gaussian process},
\newblock \bibinfo{journal}{Journal of Engineering Mechanics} \bibinfo{volume}{147} (\bibinfo{year}{2021}) \bibinfo{pages}{04021020}.
%Type = Article
\bibitem[{Miguel et~al.(2023)Miguel, Lopez, Carvalho, and Beck}]{miguel2023performance}
\bibinfo{author}{L.~F.~F. Miguel}, \bibinfo{author}{R.~H. Lopez}, \bibinfo{author}{H.~Carvalho}, \bibinfo{author}{A.~T. Beck},
\newblock \bibinfo{title}{Performance-based optimization of nonlinear friction-folded ptmds of structures subjected to stochastic excitation},
\newblock \bibinfo{journal}{Mechanical Systems and Signal Processing} \bibinfo{volume}{195} (\bibinfo{year}{2023}) \bibinfo{pages}{110291}.
%Type = Article
\bibitem[{Miguel et~al.(2024)Miguel, Elias, and Beck}]{miguel2024reliability}
\bibinfo{author}{L.~F.~F. Miguel}, \bibinfo{author}{S.~Elias}, \bibinfo{author}{A.~T. Beck},
\newblock \bibinfo{title}{Reliability-based optimization of supported pendulum tmds’ nonlinear track shape using pad{\'e } approximants},
\newblock \bibinfo{journal}{Engineering Structures} \bibinfo{volume}{306} (\bibinfo{year}{2024}) \bibinfo{pages}{117861}.
%Type = Article
\bibitem[{dos Santos et~al.(2024)dos Santos, Beck, and Lopez}]{dos2024sequential}
\bibinfo{author}{K.~R. dos Santos}, \bibinfo{author}{A.~T. Beck}, \bibinfo{author}{R.~H. Lopez},
\newblock \bibinfo{title}{Sequential simulated annealing for life-cycle optimization of nonlinear stochastic systems via arbitrary polynomial chaos expansion},
\newblock \bibinfo{journal}{Engineering Structures} \bibinfo{volume}{304} (\bibinfo{year}{2024}) \bibinfo{pages}{117675}.
%Type = Article
\bibitem[{Gu et~al.(2023)Gu, Han, Guo, Guo, Gao, and Liu}]{GU2023103514}
\bibinfo{author}{D.~Gu}, \bibinfo{author}{W.~Han}, \bibinfo{author}{J.~Guo}, \bibinfo{author}{H.~Guo}, \bibinfo{author}{S.~Gao}, \bibinfo{author}{X.~Liu},
\newblock \bibinfo{title}{A kriging-based adaptive adding point strategy for structural reliability analysis},
\newblock \bibinfo{journal}{Probabilistic Engineering Mechanics} \bibinfo{volume}{74} (\bibinfo{year}{2023}) \bibinfo{pages}{103514}.
%Type = Article
\bibitem[{Misra and Bocchini(2025)}]{misra2025metaimnet}
\bibinfo{author}{S.~Misra}, \bibinfo{author}{P.~Bocchini},
\newblock \bibinfo{title}{Metaimnet: A physics-informed neural network architecture for surrogate response and fragility modeling of structures subjected to time-varying hazard loads},
\newblock \bibinfo{journal}{Structural Safety}  (\bibinfo{year}{2025}) \bibinfo{pages}{102650}.
%Type = Article
\bibitem[{Gidaris and Taflanidis(2015)}]{gidaris2015performance}
\bibinfo{author}{I.~Gidaris}, \bibinfo{author}{A.~A. Taflanidis},
\newblock \bibinfo{title}{Performance assessment and optimization of fluid viscous dampers through life-cycle cost criteria and comparison to alternative design approaches},
\newblock \bibinfo{journal}{Bulletin of Earthquake Engineering} \bibinfo{volume}{13} (\bibinfo{year}{2015}) \bibinfo{pages}{1003--1028}.
%Type = Article
\bibitem[{Gidaris et~al.(2015)Gidaris, Taflanidis, and Mavroeidis}]{gidaris2015kriging}
\bibinfo{author}{I.~Gidaris}, \bibinfo{author}{A.~A. Taflanidis}, \bibinfo{author}{G.~P. Mavroeidis},
\newblock \bibinfo{title}{Kriging metamodeling in seismic risk assessment based on stochastic ground motion models},
\newblock \bibinfo{journal}{Earthquake Engineering \& Structural Dynamics} \bibinfo{volume}{44} (\bibinfo{year}{2015}) \bibinfo{pages}{2377--2399}.
%Type = Article
\bibitem[{Kyprioti and Taflanidis(2021)}]{kyprioti2021kriging}
\bibinfo{author}{A.~P. Kyprioti}, \bibinfo{author}{A.~A. Taflanidis},
\newblock \bibinfo{title}{Kriging metamodeling for seismic response distribution estimation},
\newblock \bibinfo{journal}{Earthquake Engineering \& Structural Dynamics} \bibinfo{volume}{50} (\bibinfo{year}{2021}) \bibinfo{pages}{3550--3576}.
%Type = Article
\bibitem[{Kim et~al.(2024)Kim, Yi, and Song}]{kim2024active}
\bibinfo{author}{J.~Kim}, \bibinfo{author}{S.-r. Yi}, \bibinfo{author}{J.~Song},
\newblock \bibinfo{title}{Active learning-based optimization of structures under stochastic excitations with first-passage probability constraints},
\newblock \bibinfo{journal}{Engineering Structures} \bibinfo{volume}{307} (\bibinfo{year}{2024}) \bibinfo{pages}{117873}.
%Type = Article
\bibitem[{Jerez et~al.(2022)Jerez, Jensen, and Beer}]{jerez2022reliability}
\bibinfo{author}{D.~Jerez}, \bibinfo{author}{H.~Jensen}, \bibinfo{author}{M.~Beer},
\newblock \bibinfo{title}{Reliability-based design optimization of structural systems under stochastic excitation: An overview},
\newblock \bibinfo{journal}{Mechanical Systems and Signal Processing} \bibinfo{volume}{166} (\bibinfo{year}{2022}) \bibinfo{pages}{108397}.
%Type = Article
\bibitem[{Zhu et~al.(2023)Zhu, Broccardo, and Sudret}]{zhu2023seismic}
\bibinfo{author}{X.~Zhu}, \bibinfo{author}{M.~Broccardo}, \bibinfo{author}{B.~Sudret},
\newblock \bibinfo{title}{Seismic fragility analysis using stochastic polynomial chaos expansions},
\newblock \bibinfo{journal}{Probabilistic Engineering Mechanics} \bibinfo{volume}{72} (\bibinfo{year}{2023}) \bibinfo{pages}{103413}.
%Type = Article
\bibitem[{Yi and Taflanidis(2024)}]{yi2024stochastic}
\bibinfo{author}{S.-r. Yi}, \bibinfo{author}{A.~A. Taflanidis},
\newblock \bibinfo{title}{Stochastic emulation with enhanced partial-and no-replication strategies for seismic response distribution estimation},
\newblock \bibinfo{journal}{Earthquake Engineering \& Structural Dynamics} \bibinfo{volume}{53} (\bibinfo{year}{2024}) \bibinfo{pages}{2354--2381}.
%Type = Article
\bibitem[{Yi and Taflanidis(2025)}]{yi2025multi}
\bibinfo{author}{S.-r. Yi}, \bibinfo{author}{A.~A. Taflanidis},
\newblock \bibinfo{title}{Multi-output stochastic emulation with applications to seismic response correlation estimation},
\newblock \bibinfo{journal}{Structural Safety}  (\bibinfo{year}{2025}) \bibinfo{pages}{102578}.
%Type = Article
\bibitem[{Kim and Wang(2025)}]{kim2025uncertainty}
\bibinfo{author}{J.~Kim}, \bibinfo{author}{Z.~Wang},
\newblock \bibinfo{title}{Uncertainty quantification for seismic response using dimensionality reduction-based stochastic simulator},
\newblock \bibinfo{journal}{Earthquake Engineering \& Structural Dynamics} \bibinfo{volume}{54} (\bibinfo{year}{2025}) \bibinfo{pages}{471--490}.
%Type = Article
\bibitem[{Zhu and Sudret(2020)}]{zhu2020replication}
\bibinfo{author}{X.~Zhu}, \bibinfo{author}{B.~Sudret},
\newblock \bibinfo{title}{Replication-based emulation of the response distribution of stochastic simulators using generalized lambda distributions},
\newblock \bibinfo{journal}{International Journal for Uncertainty Quantification} \bibinfo{volume}{10} (\bibinfo{year}{2020}).
%Type = Article
\bibitem[{Kroetz et~al.(2026)Kroetz, Beck, Costa, Macedo, Marelli, and Sudret}]{KROETZ2026122515}
\bibinfo{author}{H.~Kroetz}, \bibinfo{author}{A.~Beck}, \bibinfo{author}{L.~Costa}, \bibinfo{author}{F.~Macedo}, \bibinfo{author}{S.~Marelli}, \bibinfo{author}{B.~Sudret},
\newblock \bibinfo{title}{Fragility analysis of structures under non-synoptic winds using stochastic emulators},
\newblock \bibinfo{journal}{Engineering Structures} \bibinfo{volume}{357} (\bibinfo{year}{2026}) \bibinfo{pages}{122515}.
%Type = Article
\bibitem[{Macedo et~al.(2026)Macedo, Beck, Giannoukou, Costa, Kroetz, {Fadel Miguel}, Marelli, and Sudret}]{MACEDO2026103971}
\bibinfo{author}{F.~C. Macedo}, \bibinfo{author}{A.~T. Beck}, \bibinfo{author}{K.~Giannoukou}, \bibinfo{author}{L.~G. Costa}, \bibinfo{author}{H.~M. Kroetz}, \bibinfo{author}{L.~F. {Fadel Miguel}}, \bibinfo{author}{S.~Marelli}, \bibinfo{author}{B.~Sudret},
\newblock \bibinfo{title}{Fragility assessment of transmission line supports under non-synoptic wind loads using multi-fidelity stochastic emulators},
\newblock \bibinfo{journal}{Probabilistic Engineering Mechanics} \bibinfo{volume}{85} (\bibinfo{year}{2026}) \bibinfo{pages}{103971}.
%Type = Article
\bibitem[{Zhu and Sudret(2021{\natexlab{a}})}]{zhu2021emulation}
\bibinfo{author}{X.~Zhu}, \bibinfo{author}{B.~Sudret},
\newblock \bibinfo{title}{Emulation of stochastic simulators using generalized lambda models},
\newblock \bibinfo{journal}{SIAM/ASA Journal on Uncertainty Quantification} \bibinfo{volume}{9} (\bibinfo{year}{2021}{\natexlab{a}}) \bibinfo{pages}{1345--1380}.
%Type = Article
\bibitem[{Zhu and Sudret(2021{\natexlab{b}})}]{zhu2021global}
\bibinfo{author}{X.~Zhu}, \bibinfo{author}{B.~Sudret},
\newblock \bibinfo{title}{Global sensitivity analysis for stochastic simulators based on generalized lambda surrogate models},
\newblock \bibinfo{journal}{Reliability Engineering \& System Safety} \bibinfo{volume}{214} (\bibinfo{year}{2021}{\natexlab{b}}) \bibinfo{pages}{107815}.
%Type = Article
\bibitem[{Zhu and Sudret(2023)}]{zhu2023stochastic}
\bibinfo{author}{X.~Zhu}, \bibinfo{author}{B.~Sudret},
\newblock \bibinfo{title}{Stochastic polynomial chaos expansions to emulate stochastic simulators},
\newblock \bibinfo{journal}{International Journal for Uncertainty Quantification} \bibinfo{volume}{13} (\bibinfo{year}{2023}).
%Type = Article
\bibitem[{Pires et~al.(2025{\natexlab{a}})Pires, Moustapha, Marelli, and Sudret}]{pires2025reliability}
\bibinfo{author}{A.~V. Pires}, \bibinfo{author}{M.~Moustapha}, \bibinfo{author}{S.~Marelli}, \bibinfo{author}{B.~Sudret},
\newblock \bibinfo{title}{Reliability analysis for data-driven noisy models using active learning},
\newblock \bibinfo{journal}{Structural Safety} \bibinfo{volume}{112} (\bibinfo{year}{2025}{\natexlab{a}}) \bibinfo{pages}{102543}.
%Type = Article
\bibitem[{Pires et~al.(2025{\natexlab{b}})Pires, Moustapha, Marelli, and Sudret}]{pires2025alspce}
\bibinfo{author}{A.~V. Pires}, \bibinfo{author}{M.~Moustapha}, \bibinfo{author}{S.~Marelli}, \bibinfo{author}{B.~Sudret},
\newblock \bibinfo{title}{Reliability analysis for nondeterministic models using stochastic polynomial chaos expansion and active learning},
\newblock \bibinfo{journal}{Structural Safety (submitted)}  (\bibinfo{year}{2025}{\natexlab{b}}).
%Type = Article
\bibitem[{Moustapha and Sudret(2026)}]{moustapha2026high}
\bibinfo{author}{M.~Moustapha}, \bibinfo{author}{B.~Sudret},
\newblock \bibinfo{title}{High-dimensional reliability-based design optimization using stochastic emulators},
\newblock \bibinfo{journal}{arXiv preprint arXiv:2604.05759}  (\bibinfo{year}{2026}).
%Type = Article
\bibitem[{Melchers(1989)}]{melchers1989importance}
\bibinfo{author}{R.~Melchers},
\newblock \bibinfo{title}{Importance sampling in structural systems},
\newblock \bibinfo{journal}{Structural safety} \bibinfo{volume}{6} (\bibinfo{year}{1989}) \bibinfo{pages}{3--10}.
%Type = Article
\bibitem[{Au and Beck(2003)}]{au2003important}
\bibinfo{author}{S.-K. Au}, \bibinfo{author}{J.~L. Beck},
\newblock \bibinfo{title}{Important sampling in high dimensions},
\newblock \bibinfo{journal}{Structural Safety} \bibinfo{volume}{25} (\bibinfo{year}{2003}) \bibinfo{pages}{139--163}.
%Type = Article
\bibitem[{Au and Beck(2001)}]{au2001estimation}
\bibinfo{author}{S.-K. Au}, \bibinfo{author}{J.~L. Beck},
\newblock \bibinfo{title}{Estimation of small failure probabilities in high dimensions by subset simulation},
\newblock \bibinfo{journal}{Probabilistic Engineering Mechanics} \bibinfo{volume}{16} (\bibinfo{year}{2001}) \bibinfo{pages}{263--277}.
%Type = Article
\bibitem[{Au and Beck(2003)}]{au2003subset}
\bibinfo{author}{S.-K. Au}, \bibinfo{author}{J.~L. Beck},
\newblock \bibinfo{title}{Subset simulation and its application to seismic risk based on dynamic analysis},
\newblock \bibinfo{journal}{Journal of Engineering Mechanics} \bibinfo{volume}{129} (\bibinfo{year}{2003}) \bibinfo{pages}{901--917}.
%Type = Article
\bibitem[{Wang et~al.(2019)Wang, Broccardo, and Song}]{wang2019hamiltonian}
\bibinfo{author}{Z.~Wang}, \bibinfo{author}{M.~Broccardo}, \bibinfo{author}{J.~Song},
\newblock \bibinfo{title}{Hamiltonian monte carlo methods for subset simulation in reliability analysis},
\newblock \bibinfo{journal}{Structural Safety} \bibinfo{volume}{76} (\bibinfo{year}{2019}) \bibinfo{pages}{51--67}.
%Type = Article
\bibitem[{Chen et~al.(2022)Chen, Wang, Broccardo, and Song}]{chen2022riemannian}
\bibinfo{author}{W.~Chen}, \bibinfo{author}{Z.~Wang}, \bibinfo{author}{M.~Broccardo}, \bibinfo{author}{J.~Song},
\newblock \bibinfo{title}{Riemannian manifold hamiltonian monte carlo based subset simulation for reliability analysis in non-gaussian space},
\newblock \bibinfo{journal}{Structural Safety} \bibinfo{volume}{94} (\bibinfo{year}{2022}) \bibinfo{pages}{102134}.
%Type = Article
\bibitem[{Arunachalam and Spence(2023)}]{ARUNACHALAM2023102310}
\bibinfo{author}{S.~Arunachalam}, \bibinfo{author}{S.~M.~J. Spence},
\newblock \bibinfo{title}{An efficient stratified sampling scheme for the simultaneous estimation of small failure probabilities in wind engineering applications},
\newblock \bibinfo{journal}{Structural Safety} \bibinfo{volume}{101} (\bibinfo{year}{2023}) \bibinfo{pages}{102310}.
%Type = Article
\bibitem[{Xian and Wang(2024)}]{xian2024relaxation}
\bibinfo{author}{J.~Xian}, \bibinfo{author}{Z.~Wang},
\newblock \bibinfo{title}{Relaxation-based importance sampling for structural reliability analysis},
\newblock \bibinfo{journal}{Structural Safety} \bibinfo{volume}{106} (\bibinfo{year}{2024}) \bibinfo{pages}{102393}.
%Type = Article
\bibitem[{Lee et~al.(2025)Lee, Wang, and Song}]{LEE2025110947}
\bibinfo{author}{D.~Lee}, \bibinfo{author}{Z.~Wang}, \bibinfo{author}{J.~Song},
\newblock \bibinfo{title}{Efficient seismic reliability and fragility analysis of lifeline networks using subset simulation},
\newblock \bibinfo{journal}{Reliability Engineering \& System Safety} \bibinfo{volume}{260} (\bibinfo{year}{2025}) \bibinfo{pages}{110947}.
%Type = Article
\bibitem[{Dubourg et~al.(2013)Dubourg, Sudret, and Deheeger}]{DUBOURG201347}
\bibinfo{author}{V.~Dubourg}, \bibinfo{author}{B.~Sudret}, \bibinfo{author}{F.~Deheeger},
\newblock \bibinfo{title}{Metamodel-based importance sampling for structural reliability analysis},
\newblock \bibinfo{journal}{Probabilistic Engineering Mechanics} \bibinfo{volume}{33} (\bibinfo{year}{2013}) \bibinfo{pages}{47--57}.
%Type = Article
\bibitem[{Zhu et~al.(2020)Zhu, Lu, and Yun}]{ZHU2020106644}
\bibinfo{author}{X.~Zhu}, \bibinfo{author}{Z.~Lu}, \bibinfo{author}{W.~Yun},
\newblock \bibinfo{title}{An efficient method for estimating failure probability of the structure with multiple implicit failure domains by combining meta-is with is-ak},
\newblock \bibinfo{journal}{Reliability Engineering \& System Safety} \bibinfo{volume}{193} (\bibinfo{year}{2020}) \bibinfo{pages}{106644}.
%Type = Article
\bibitem[{Xiao et~al.(2020)Xiao, Zhan, and Yuan}]{XIAO2020113336}
\bibinfo{author}{N.-C. Xiao}, \bibinfo{author}{H.~Zhan}, \bibinfo{author}{K.~Yuan},
\newblock \bibinfo{title}{A new reliability method for small failure probability problems by combining the adaptive importance sampling and surrogate models},
\newblock \bibinfo{journal}{Computer Methods in Applied Mechanics and Engineering} \bibinfo{volume}{372} (\bibinfo{year}{2020}) \bibinfo{pages}{113336}.
%Type = Article
\bibitem[{Yu and Lu(2023)}]{YU2023108722}
\bibinfo{author}{T.~Yu}, \bibinfo{author}{Z.~Lu},
\newblock \bibinfo{title}{A novel single-loop kriging importance sampling method for estimating failure probability upper bound under random-interval mixed uncertainties},
\newblock \bibinfo{journal}{Aerospace Science and Technology} \bibinfo{volume}{143} (\bibinfo{year}{2023}) \bibinfo{pages}{108722}.
%Type = Article
\bibitem[{Arunachalam and Spence(2023)}]{arunachalam2023generalized}
\bibinfo{author}{S.~Arunachalam}, \bibinfo{author}{S.~M.~J. Spence},
\newblock \bibinfo{title}{Generalized stratified sampling for efficient reliability assessment of structures against natural hazards},
\newblock \bibinfo{journal}{Journal of Engineering Mechanics} \bibinfo{volume}{149} (\bibinfo{year}{2023}) \bibinfo{pages}{04023042}.
%Type = Article
\bibitem[{Baker and Cornell(2008)}]{baker2008uncertainty}
\bibinfo{author}{J.~W. Baker}, \bibinfo{author}{C.~A. Cornell},
\newblock \bibinfo{title}{Uncertainty propagation in probabilistic seismic loss estimation},
\newblock \bibinfo{journal}{Structural Safety} \bibinfo{volume}{30} (\bibinfo{year}{2008}) \bibinfo{pages}{236--252}.
%Type = Book
\bibitem[{L{\"u}then et~al.(2026)L{\"u}then, Zhu, Marelli, and Sudret}]{luthen2024uqlab}
\bibinfo{author}{N.~L{\"u}then}, \bibinfo{author}{X.~Zhu}, \bibinfo{author}{S.~Marelli}, \bibinfo{author}{B.~Sudret}, \bibinfo{title}{UQLab user manual -- Stochastic polynomial chaos expansions ({SPCE})}, \bibinfo{publisher}{Chair of Risk, Safety and Uncertainty Quantification, ETH Zurich, Switzerland}, \bibinfo{address}{Zurich}, \bibinfo{year}{2026}. \bibinfo{note}{Report UQLab-V2.2-121}.
%Type = Article
\bibitem[{{AISC}(2010)}]{american2002seismic}
\bibinfo{author}{{AISC}},
\newblock \bibinfo{title}{{ANSI/AISC} 341-10: Seismic provisions for structural steel buildings},
\newblock \bibinfo{journal}{American Institute of Steel Construction}  (\bibinfo{year}{2010}).
%Type = Article
\bibitem[{{ASCE}(2013)}]{american2013minimum}
\bibinfo{author}{{ASCE}},
\newblock \bibinfo{title}{{ASCE}/{SEI} 7-10: Minimum design loads for buildings and other structures},
\newblock \bibinfo{journal}{American Society of Civil Engineers}  (\bibinfo{year}{2013}).
%Type = Article
\bibitem[{Uang et~al.(2004)Uang, Nakashima, and Tsai}]{uang2004research}
\bibinfo{author}{C.-M. Uang}, \bibinfo{author}{M.~Nakashima}, \bibinfo{author}{K.-C. Tsai},
\newblock \bibinfo{title}{Research and application of buckling-restrained braced frames},
\newblock \bibinfo{journal}{International Journal of Steel Structures} \bibinfo{volume}{4} (\bibinfo{year}{2004}) \bibinfo{pages}{301--313}.
%Type = Article
\bibitem[{NIST(2010)}]{nist2010evaluation}
\bibinfo{author}{NIST},
\newblock \bibinfo{title}{{NIST GCR} 10-917-8: Evaluation of the {FEMA} {P}695 methodology for quantification of building seismic performance factors},
\newblock \bibinfo{journal}{National Institute of Standards and Technology}  (\bibinfo{year}{2010}).
%Type = Article
\bibitem[{Ghasemof et~al.(2022)Ghasemof, Mirtaheri, and Mohammadi}]{ghasemof2022multi}
\bibinfo{author}{A.~Ghasemof}, \bibinfo{author}{M.~Mirtaheri}, \bibinfo{author}{R.~K. Mohammadi},
\newblock \bibinfo{title}{Multi-objective optimization for probabilistic performance-based design of buildings using fema p-58 methodology},
\newblock \bibinfo{journal}{Engineering Structures} \bibinfo{volume}{254} (\bibinfo{year}{2022}) \bibinfo{pages}{113856}.
%Type = Article
\bibitem[{Guerrero et~al.(2017)Guerrero, Ter{\'a}n-Gilmore, Ji, and Escobar}]{guerrero2017evaluation}
\bibinfo{author}{H.~Guerrero}, \bibinfo{author}{A.~Ter{\'a}n-Gilmore}, \bibinfo{author}{T.~Ji}, \bibinfo{author}{J.~A. Escobar},
\newblock \bibinfo{title}{Evaluation of the economic benefits of using buckling-restrained braces in hospital structures located in very soft soils},
\newblock \bibinfo{journal}{Engineering Structures} \bibinfo{volume}{136} (\bibinfo{year}{2017}) \bibinfo{pages}{406--419}.
%Type = Article
\bibitem[{AISC(2022)}]{design1999specification}
\bibinfo{author}{AISC},
\newblock \bibinfo{title}{Specification for structural steel buildings},
\newblock \bibinfo{journal}{ANSI/AISC}  (\bibinfo{year}{2022}).
%Type = Misc
\bibitem[{McKenna et~al.(2006)McKenna, Fenves, and Scott}]{mckenna2006opensees}
\bibinfo{author}{F.~McKenna}, \bibinfo{author}{G.~Fenves}, \bibinfo{author}{M.~Scott}, \bibinfo{title}{Opensees: Open system for earthquake engineering simulation. pacific earthquake engineering research center, uc berkeley, ca}, \bibinfo{year}{2006}.
%Type = Book
\bibitem[{FILIPPOU et~al.(1983)FILIPPOU, POPOV, and BERTERO}]{filippou1983effects}
\bibinfo{author}{F.~C. FILIPPOU}, \bibinfo{author}{E.~P. POPOV}, \bibinfo{author}{V.~V. BERTERO}, \bibinfo{title}{Effects of bond deterioration on hysteretic behavior of reinforced concrete joints}, \bibinfo{publisher}{Earthquake Engineering Research Center, University of California, College of Engineering}, \bibinfo{address}{Berkeley}, \bibinfo{year}{1983}. \bibinfo{note}{(Report n\textsuperscript{o} UCB/EERC/19, August 1983)}.
%Type = Book
\bibitem[{Holland(1992)}]{holland1992adaptation}
\bibinfo{author}{J.~H. Holland}, \bibinfo{title}{Adaptation in natural and artificial systems: an introductory analysis with applications to biology, control, and artificial intelligence}, \bibinfo{publisher}{MIT Press}, \bibinfo{address}{Cambridge}, \bibinfo{year}{1992}.
%Type = Article
\bibitem[{Vetter and Taflanidis(2012)}]{vetter2012global}
\bibinfo{author}{C.~Vetter}, \bibinfo{author}{A.~A. Taflanidis},
\newblock \bibinfo{title}{Global sensitivity analysis for stochastic ground motion modeling in seismic-risk assessment},
\newblock \bibinfo{journal}{Soil Dynamics and Earthquake Engineering} \bibinfo{volume}{38} (\bibinfo{year}{2012}) \bibinfo{pages}{128--143}.
%Type = Article
\bibitem[{Atkinson and Silva(2000)}]{atkinson2000stochastic}
\bibinfo{author}{G.~M. Atkinson}, \bibinfo{author}{W.~Silva},
\newblock \bibinfo{title}{Stochastic modeling of california ground motions},
\newblock \bibinfo{journal}{Bulletin of the Seismological Society of America} \bibinfo{volume}{90} (\bibinfo{year}{2000}) \bibinfo{pages}{255--274}.
%Type = Article
\bibitem[{Boore(2003)}]{boore2003simulation}
\bibinfo{author}{D.~M. Boore},
\newblock \bibinfo{title}{Simulation of ground motion using the stochastic method},
\newblock \bibinfo{journal}{Pure and applied geophysics} \bibinfo{volume}{160} (\bibinfo{year}{2003}) \bibinfo{pages}{635--676}.
%Type = Article
\bibitem[{Boore and Joyner(1997)}]{boore1997site}
\bibinfo{author}{D.~M. Boore}, \bibinfo{author}{W.~B. Joyner},
\newblock \bibinfo{title}{Site amplifications for generic rock sites},
\newblock \bibinfo{journal}{Bulletin of the seismological society of America} \bibinfo{volume}{87} (\bibinfo{year}{1997}) \bibinfo{pages}{327--341}.

\end{thebibliography}





% \end{thebibliography}
\end{document}